\documentclass[journal]{IEEEtran}
\usepackage{bibunits}
\usepackage{amsmath,amsfonts}
\usepackage{algorithmic}
\usepackage{algorithm}
\usepackage{array}
\usepackage[caption=false,font=normalsize,labelfont=sf,textfont=sf]{subfig}
\usepackage{textcomp}
\usepackage{stfloats}
\usepackage{url}
\usepackage{verbatim}
\usepackage{graphicx}
\usepackage{cite}
\usepackage{tablefootnote}
\usepackage{siunitx}
\usepackage{booktabs}
\usepackage{amsfonts}
\usepackage{amssymb}
\usepackage{amsmath}
\usepackage{upgreek}

\usepackage{wasysym}

\usepackage{multirow}
\usepackage{subfiles}
\usepackage{color}
\usepackage{xr}
\defaultbibliographystyle{IEEEtran}

\newcommand{\specialcell}[2][c]{%
\begin{tabular}[#1]{@{}c@{}}#2\end{tabular}}

\usepackage{tikz}
\newcommand*{\mycirc}[1]{%
  \tikz[baseline=(char.base)]{
    \node[draw,circle,inner sep=1pt] (char) {#1};
  }%
}

\begin{document}

\title{VS-Splat: Voxel-Selective feed-forward Gaussian Splatting for end-to-end 3D object reconstruction from sparse-views}

\author{
Yunsu Jeong\textsuperscript{*},
Hyuk Heo\textsuperscript{*},
Youngsang Kwak,
Jaehwa Kwak,
and Il Yong Chun, \IEEEmembership{Member, IEEE}
\thanks{
\textsuperscript{*}These authors contributed equally.
The work was supported in part
by NRF Grants RS-2026-25480389 and RS2023-00213455 funded by MSIT, 
IITP Grants
RS-2026-25615334 (AI Star Fellowship Support Program) and
RS-2019-II190421 (AI Graduate School Support Program (Sungkyunkwan
University)) funded by MSIT,
the BK21 FOUR Project,
the Digital Therapeutics Development and Demonstration Support Program Grant
H0601-24-1023 funded by MSIT and NIPA, and
KIAT Grant RS-2024-00418086 (HRD Program
for Industrial Innovation) funded by MOTIE. (Corresponding author: Il Yong Chun.)}
\thanks{Yunsu Jeong is with the Department of Electrical and Computer Engineering (ECE), Sungkyunkwan University (SKKU), Suwon 16419, South Korea (e-mail: intyeger@g.skku.edu).
}
\thanks{
 Hyuk Heo is with the Department of Display Convergence Engineering (DCE), SKKU, Suwon 16419, South Korea (e-mail:mongyong2@g.skku.edu).
}
\thanks{
 Youngsang Kwak and Jaehwa Kwak are with AiM Future, Seoul 06804, South Korea (email: \{youngsang.kwak, jaehwa.kwak\}@aimfuture.ai)
}
\thanks{
Il Yong Chun is with the Departments of ECE, Artificial Intelligence, Advanced Display Engineering,  Semiconductor Convergence Engineering, and DCE, SKKU, Suwon
16419, South Korea (e-mail:
iychun@skku.edu).
}

}

\maketitle
\begin{bibunit}
\begin{abstract}
Feed-forward Gaussian splatting models have demonstrated remarkable effectiveness in reconstructing three-dimensional (3D) objects from a few two-dimensional (2D) images, even if they are unseen.
As existing methods typically predict Gaussian primitives uniformly across the 3D space, most primitives are placed in non-object regions.
This may hinder the representation of fine object details.
This paper proposes a Voxel-Selective Gaussian Splatting model ({\bfseries VS-Splat}), a new end-to-end feed-forward Gaussian splatting framework that predicts many primitives only within selected voxels that are likely to belong to an object, without 3D structural supervision.
To achieve this, we propose a new learnable voxel selection approach that identifies object-centric voxels only with 2D rendering supervision.
Our sparse-view rendering experiments with three benchmark datasets show that proposed VS-Splat outperforms several state-of-the-art methods.
We further demonstrate its effectiveness as a backbone for an existing densification method and show that an optional extension improves its robustness to inaccurate camera pose estimates.
\end{abstract}


\section{Introduction}
\IEEEPARstart{A}{}wide range of industries including games, virtual reality, augmented reality, medical imaging, and film production increasingly require high-quality three-dimensional (3D) data \cite{ldm3dvr,dx2ct, mobile3dscanner, scalegs, magic3d}. 
While two-dimensional (2D) images are readily available at scale, high-quality 3D data remain relatively scarce due to the difficulty and cost of reconstruction.  
In particular, 3D object reconstruction has been extensively studied as a fundamental task for downstream multimedia applications.
Recent advances in Neural Radiance Fields (NeRF) \cite{nerf} and 3D Gaussian Splatting (3DGS) \cite{3dgs} have significantly improved 3D object reconstruction by learning effective 3D representations from multi-view images of an isolated object \cite{sugar} or an object segmented from a scene \cite{sa3d}.
Although 3DGS enables real-time rendering compared to the computationally expensive volumetric rendering of NeRF, 3DGS \cite{3dgs} and its variants \cite{sa3d,sugar} 
still require iterative optimization and typically rely on densely captured multi-view images and iterative optimization to reconstruct a \emph{single} 3D object.

To overcome these limitations, a new class of generalizable feed-forward Gaussian splatting (GS) models has emerged.
Such models leverage a 3D prior learned from large-scale multi-view data, enabling the direct generation of Gaussian representations from a single image or a few sparse-view images via a single forward pass, without per-object optimization.

\begin{figure}[!t]
    \small
    \centering
    \setlength{\tabcolsep}{1pt}
    \begin{tabular}{c|c}
        \raisebox{0mm}{\includegraphics[width=0.20\linewidth]{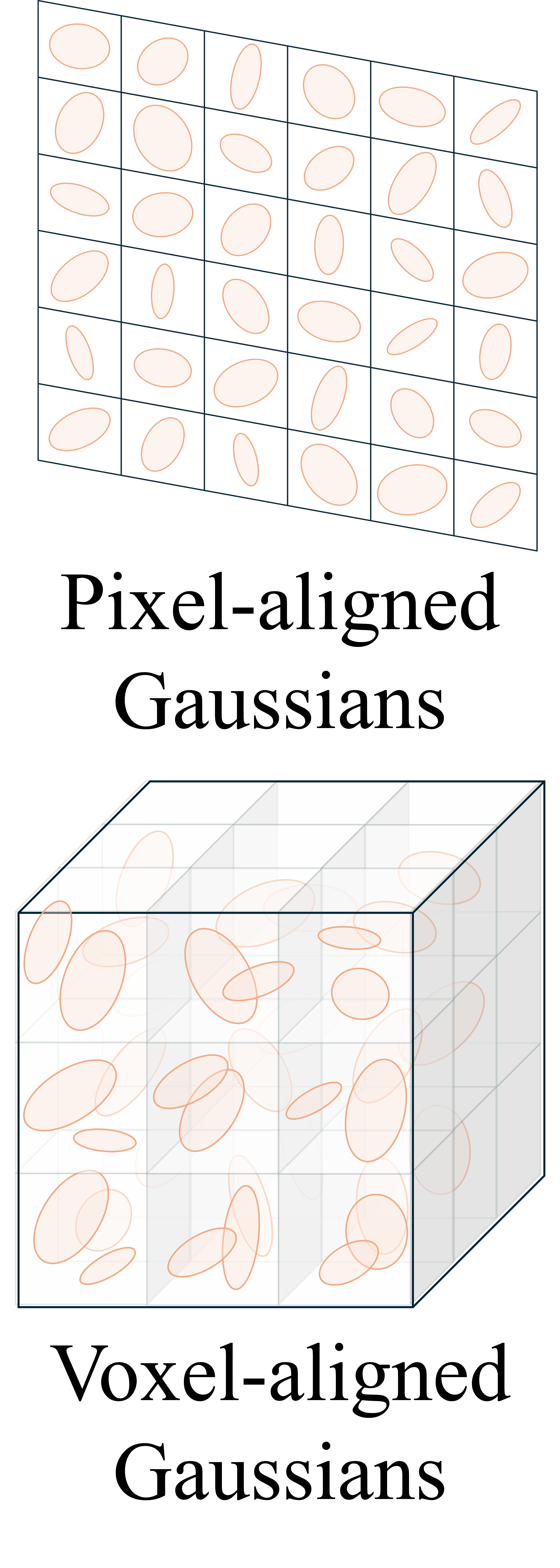}} &
        \raisebox{3mm}{\includegraphics[width=0.78\linewidth]{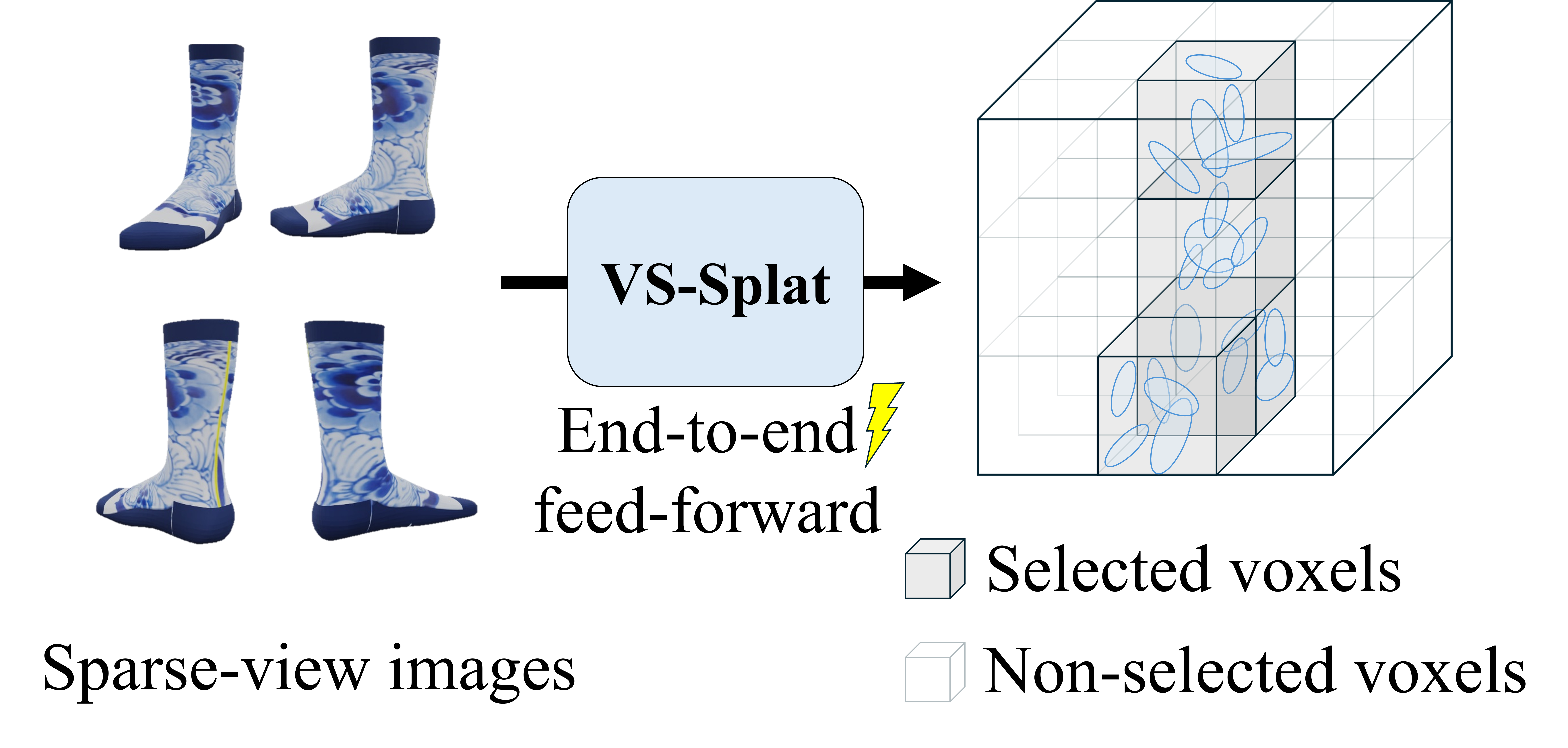}} \\
        \specialcell{(a) Existing~ \\ methods} &
        {(b) \textbf{VS-Splat (ours)}}
    \end{tabular}

    \caption{Overview of the proposed framework vs.~existing E2E feed-forward object reconstruction approaches in generalizable Gaussian splatting under a \emph{sparse-view} setup.
    (a) Existing methods predict a fixed, small number of Gaussian primitives at every pixel or voxel, regardless of object occupancy.  
    (b) The proposed {\bfseries VS-Splat} framework first selects voxels that are likely to belong to an object, and then predicts many Gaussian primitives at each selected voxel, all within a single end-to-end feed-forward pass. Its training does \emph{not} require 3D structural supervision.}
    \vspace{-1pc}
    
    \label{fig:overview}
\end{figure}

These feed-forward GS approaches typically adopt either a pixel-aligned strategy \cite{splatter-image} that places Gaussian primitives within 2D pixel grid cells, or a voxel-aligned strategy \cite{lara,gs-rgbn} that places them within 3D voxel grid cells.
By placing Gaussian primitives within each pixel or voxel, the mapping from 2D image(s) to Gaussian representations becomes more explicit and efficient.
However, as spatial resolution of the underlying grid increases, the number of grid cells grows quadratically in the pixel-aligned strategy and cubically in the voxel-aligned strategy, thereby restricting each cell to contain only one or two Gaussian primitives for efficient novel view synthesis.
This limitation becomes particularly inefficient for 3D objects surrounded by large empty regions,
as many Gaussian primitives are allocated to uninformative empty regions rather than to an object itself.

Another stream of generalizable GS explores 3D geometry-aware GS research investigates
3D geometry-aware GS models trained with a large-scale 3D geometry dataset. 
These geometry-aware methods \cite{geolrm,pm-loss} adopt a multi-stage pipeline: a first network estimates the 3D object geometry and predicts the positions of Gaussian primitives, followed by a separate network that predicts their parameters. 
Yet, these methods are \emph{not} end-to-end (E2E)  with respect to rendering fidelity. 
As a result, they are inherently prone to error propagation, since each stage is trained independently – either with supervision from 3D geometries (in the first network) or from 2D images (in the second network).

We propose a Voxel-Selective Gaussian Splatting model, called \textbf{VS-Splat}, a new E2E feed-forward GS framework for novel view synthesis from a few 2D views \emph{without} 3D structural supervision.
VS-Splat is built around a central E2E design principle: 
rendering supervision determines not only the parameters of Gaussian primitives but also their spatial allocation.
Accordingly, VS-Splat predicts many Gaussian primitives only in selected voxels likely to belong to an object.
Fig.~\ref{fig:overview} highlights this distinction between VS-Splat and existing E2E feed-forward GS approaches for sparse-view 3D object reconstruction.

Specifically, the proposed VS-Splat framework employs a coarse-to-fine (C2F) approach strategically allocates Gaussian primitives to object-centric regions. 
The coarse stage of VS-Splat estimates the coarse voxelized structure of an object, and predicts a single coarse Gaussian primitive at each voxel. 
Next, we select voxels that are likely to belong to an object using the coarse, global structural feature. 
We achieve this \emph{without} any 3D structural supervision, by casting 3D structure estimation of an object as a voxel-wise binary classification task, where the outputs consist of a confidence score map and a voxel selection mask identifying target voxels for dense Gaussian primitive placement. 
For E2E learning, we propose a new learnable voxel selection approach that integrates the resulting selection confidence scores into the GS rendering process. 
To enhance the representation of object details, 
the fine stage of VS-Splat predicts many additional fine-grained Gaussian primitives near selected coarse ones. 
We propose a network that enriches coarse structural feature by incorporating fine local details at the selected voxels. 

Our contributions are summarized as follows:

\begin{itemize}
    \item We propose a new E2E feed-forward Gaussian splatting model VS-Splat that selects voxels likely to belong to an object and then predicts many fine-grained Gaussian primitives at those voxels.
    
    \item We propose a new learnable voxel selection approach that circumvents explicit 3D structure estimation, relying on 2D rendering supervision for training.

    \item To predict many fine-grained Gaussian primitives, we propose a selective feature refinement network that refines coarse, global features by integrating fine, local geometric features, at selected object-centric voxels.

    \item Our sparse-view rendering experiments with the three benchmark datasets, large-scale GObjaverse \cite{gobjaverse}, Google Scanned Objects (GSO) \cite{gso}, and Common Objects in 3D (CO3D) \cite{co3d}, show that VS-Splat outperforms existing state-of-the-art (SOTA) methods. 
    
    \item We further demonstrate the extensibility of VS-Splat in two settings.
    It consistently improves rendering performance when used in place of the original backbone of an existing densification method \cite{GenerativeDensification}.
    We also introduce an optional extension that improves its robustness to noisy camera pose estimates on in-the-wild data. 
\end{itemize}

\begin{figure*}[!ht]
    \centering
    \includegraphics[width=1.0\linewidth]{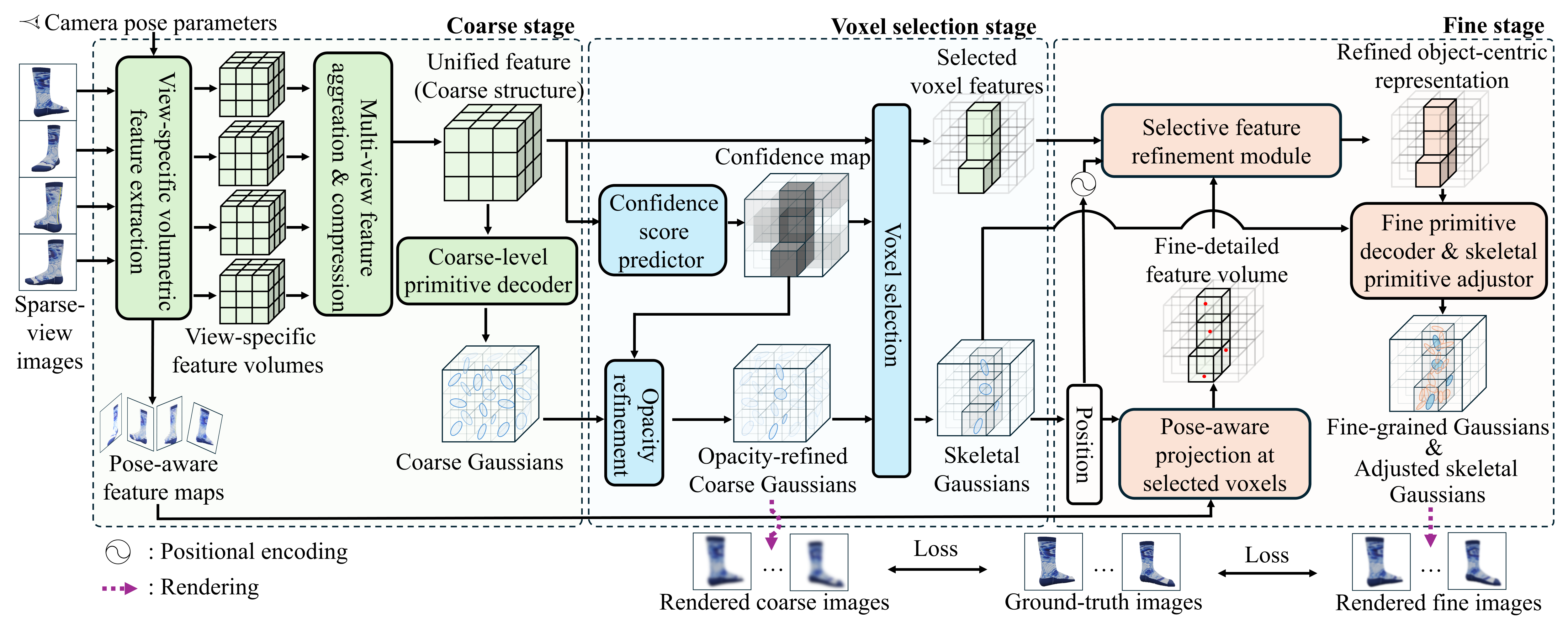}
    
    \vspace{-1pc}
    \caption{
    Overall architecture of proposed VS-Splat. 
    It consists of three stages: coarse, voxel selection, and fine.}   
    \vspace{-0.75pc}
    
    \label{fig:architecture}
\end{figure*}

\section{Related works}
\subsection{Feed-forward GS methods}

Recent feed-forward GS methods reconstruct a 3D representation of a scene or an object by predicting Gaussian primitives in 3D space from one or more input images.

\subsubsection{Scene reconstruction methods}

Scene reconstruction methods mainly adopt a pixel-aligned strategy \cite{pixelsplat,mvsplat,depthsplat,splatweaver}, 
in which Gaussian predictions are indexed by a regular 2D grid associated with each input view.
In methods with known camera parameters, Gaussian centers are typically obtained by predicting a depth or depth distribution at each grid location and locating the centers along the corresponding camera rays \cite{pixelsplat,mvsplat,depthsplat}.
For methods that do not assume known camera parameters, Gaussian centers may instead be directly regressed from pixel-wise features \cite{splatweaver}.

\subsubsection{Object reconstruction methods}

Object reconstruction methods broadly adopt three Gaussian prediction strategies: 
pixel-aligned \cite{splatter-image, lgm, gs-lrm}, 
voxel-aligned \cite{gs-rgbn, lara, GenerativeDensification}, and 
geometry-aware \cite{tgs, geolrm}.
The pixel- and voxel-aligned strategies are illustrated in Fig.~\ref{fig:overview}(a).

Pixel-aligned methods organize Gaussian predictions on a dense 2D grid associated with each input view and predict one or a few primitives at every grid location.
Because each grid location provides only a fixed number of primitive prediction slots, 
a dense grid is used to provide sufficient representational capacity for fine image details.
Such dense prediction remains computationally feasible because the number of grid locations grows quadratically with grid resolution.
Depending on the method, Gaussian centers are either constrained to the corresponding camera rays \cite{splatter-image, gs-lrm} or directly regressed in 3D \cite{lgm}.

Voxel-aligned methods instead define a regular 3D voxel grid and predict one or a few Gaussian primitives within every voxel.
Since the number of voxels grows cubically with grid resolution, increasing the spatial resolution is substantially more expensive than for a 2D pixel grid.
Consequently, voxel-aligned methods typically operate on a lower grid resolution, and their uniform allocation also assigns primitive prediction slots to empty voxels.
As illustrated in Fig.~1(a), both pixel- and voxel-aligned methods initially allocate a fixed number of primitives over their respective grids, including background pixels or empty voxels.

Geometry-aware methods first estimate an explicit 3D structure, such as a point cloud or occupied anchors, using 3D supervision and then decode Gaussian primitives from the estimated structure using rendering-based supervision~\cite{tgs,geolrm}.
Because their Gaussian prediction depends on an intermediate 3D representation learned with explicit 3D supervision, 
these methods are \emph{not} E2E with respect to rendering supervision alone (thus falling outside the scope of the E2E comparison in Fig.~\ref{fig:overview}(a)).
In contrast, the aforementioned pixel- and voxel-aligned methods are trained E2E with rendering-based supervision.

In the \emph{sparse-view} setting, 
LGM \cite{lgm} and GS-LRM \cite{gs-lrm} adopt pixel-aligned prediction, LaRa  \cite{lara} and Generative Densification (GenDen) \cite{GenerativeDensification} adopt voxel-aligned prediction, while GeoLRM \cite{geolrm} adopts geometry-aware prediction.

\subsection{C2F approaches in 3D reconstruction}

C2F approaches have been adopted in various forms across 3D reconstruction.
Earlier methods typically realize C2F by progressively refining coarse representations into finer ones.
Point-based methods (e.g., point clouds) realize C2F by progressively increasing point density in selected regions \cite{crn}.
Volume-based methods, which represent 3D space using voxel grids storing color and opacity, realize C2F by progressively increasing grid resolution \cite{neuralrecon}.
Neural implicit representation methods realize C2F by progressively refining ray sampling around object surfaces through hierarchical sampling \cite{nerf,dsnerf, near_surf_nerf, nerfacc}.

\subsubsection{C2F approaches in feed-forward GS methods for 3D object reconstruction}

Recent feed-forward GS methods for 3D object reconstruction have explored different C2F strategies for refining Gaussian representations. 
LaRa predicts two Gaussian primitives at each voxel throughout a fixed voxel grid and selectively refines the view-dependent appearance of primitives whose opacity exceeds a predefined threshold, without generating additional primitives \cite{lara}.
For object reconstruction, GenDen augments LaRa with an add-on densification module that selects a subset of LaRa-generated Gaussian primitives based on their view-space positional gradients and generates additional fine Gaussian primitives around them \cite{GenerativeDensification}.

Unlike LaRa and GenDen, which select already predicted Gaussian primitives for appearance refinement or densification, respectively, 
the proposed VS-Splat performs selection at the voxel level before fine Gaussian generation. 
Specifically, VS-Splat predicts one coarse Gaussian primitive at each voxel and estimates a voxel-wise selection mask from the coarse volumetric features to identify voxels likely to belong to an object. 
It then generates multiple fine Gaussian primitives only within the selected voxels. 
Thus, the coarse representation determines where new fine Gaussian primitives are instantiated, rather than which existing primitives should be refined or densified.

\subsubsection{Selection strategies for Gaussian refinement and allocation}

Several feed-forward GS methods employ different criteria to select Gaussian primitives or spatial regions for refinement, densification, or adaptive primitive allocation.
LaRa uses opacity as its selection criterion and applies its fine appearance decoder only to primitives whose opacity exceeds a predefined threshold \cite{lara}.
GenDen first selects Gaussian primitives with large view-space positional gradients for densification and subsequently uses learned confidence scores to determine which generated primitives undergo further densification \cite{GenerativeDensification}.
For scene-level generalizable novel view synthesis, 
SplatWeaver uses predicted routing scores to select a cardinality expert for each pixel grid cell, thereby determining the number of Gaussian primitives generated at that cell \cite{splatweaver}.
In supervising this routing process, 
it uses pseudo-labels obtained by ranking 2D image pixels according to their frequency energy, in addition to the rendering objective.
These discrete selection operations are non-differentiable, 
and GenDen and SplatWeaver employ a straight-through estimator (STE)~\cite{ste} to train their score-based selection modules.

The proposed VS-Splat learns voxel selection in an \emph{E2E learnable} manner and allocates fine Gaussian primitives only within voxels likely to belong to an object.
Its selection module is trained solely through the rendering objective, without pseudo-labels or other auxiliary supervision.

\section{Methods}

We propose the VS-Splat framework that takes $N$ 2D images from different views $\{ \mathbf{i}_n : n = 1,\ldots, N \}$ and their corresponding camera pose parameters (as input), 
and generates 3D Gaussian representations aligned to selected voxels (as output).
Throughout this study, we focus on a sparse-view setup, i.e., $\{ \mathbf{i}_n \}$ are a few sparse-view images.
Fig.~\ref{fig:architecture} illustrates the overall architecture of VS-Splat that adopts a C2F approach:
\begin{itemize}
\item The coarse stage extracts feature maps from sparse-view images, aggregates them into a unified feature volume that captures coarse, global structure of an object.
From this feature volume, we predict a coarse Gaussian primitive at each voxel across the entire volumetric space.
See Fig.~\ref{fig:architecture}: Coarse stage.

\item Next, we select voxels that are likely to belong to an object by a new learnable voxel selection approach.
We refer to the coarse Gaussian primitives corresponding to the selected voxels as ``skeletal'' Gaussian primitives.
See Fig.~\ref{fig:architecture}: Voxel selection stage.

\item The fine stage predicts many fine-grained Gaussian primitives near a ``skeletal'' primitive at each selected voxel, via a new selective feature refinement module that refines coarse, global features by integrating fine, local geometric features, at object-centric voxels.   
See Fig.~\ref{fig:architecture}: Fine stage.

\end{itemize}

\noindent In training, we propose to use 2D supervision in both coarse and fine renderings. See the bottom part of Fig.~\ref{fig:architecture}.
In addition to the overall design, Fig.~\ref{fig:architecture} illustrates the detailed workflow of each stage.

\subsection{Backgrounds}
\label{method:backgrounds}

3DGS represents a 3D asset with a set of $K$ 3D Gaussian primitives $\{ \mathcal{G}_k: k=1,\ldots,K \}$ \cite{3dgs}.
Each Gaussian primitive $\mathcal{G}_k$ is parameterized by 
\textit{1)} a mean vector $\boldsymbol{\upmu}_k \in \mathbb{R}^3$ representing its 3D position,
\textit{2)} a covariance matrix $\mathbf{\Sigma}_k \in \mathbb{R}^{3 \times 3}$ representing its 3D shape that can be decomposed into a scale vector $\mathbf{s}_k \in \mathbb{R}^3$ and a rotation quaternion $\mathbf{r}_k \in \mathbb{R}^4$, 
\textit{3)} an opacity $o_k \in [0,1]$, and
\textit{4)} spherical harmonics coefficients $\mathbf{h}_k \in \mathbb{R}^{3S}$ representing view-dependent color, where $S$ is the number of basis functions per color channel (red, green, and blue).
These primitives are passed through the Gaussian splatting rasterization \cite{3dgs}, producing 2D images rendered from arbitrary viewpoints.

The voxel-aligned feed-forward Gaussian splatting models predict Gaussian primitives aligned to a voxel grid from sparse-view images \cite{lara,gs-rgbn}, so that $K$ becomes the number of voxels in a voxel grid.
All views share the predefined voxel grid within the normalized 3D space $[-0.5, 0.5]^3$, centered at the origin, 
where the voxel grid uses the center-coordinate system.
For the $k$th voxel, instead of directly predicting the position of its Gaussian primitive, $\boldsymbol{\upmu}_k$, the methods predict its positional offset $\mathbf{\Delta}_k \in [-r,r]^3$, where $r \in \mathbb{R}_{>0}$ is the maximum displacement, 
and add the predicted offset to its coordinate, for $k = 1,\ldots,K$.
This strategy can ensure that each primitive is placed within the corresponding voxel.
Similar to 3DGS, the primitives are rendered through rasterization.

\subsection{The coarse stage}
\label{sec:cs}

The proposed coarse stage consists of the following three core modules: 
\textit{1)} view-specific volumetric feature extraction, 
\textit{2)} multi-view feature aggregation \& compression, and 
\textit{3)} coarse primitive decoder.

\subsubsection{View-specific volumetric feature extraction.}
\label{sec:fe}

We first obtain a volumetric feature from a 2D input image at each view $\mathbf{i}_n$, for $n = 1,\ldots,N$,  
by following the approach in \cite{lara,gs-rgbn}.
As its first step, 2D feature maps are extracted from each view image, using a shared feature extractor.
For each view $n$, its feature map is then modulated using the adaptive layer normalization (AdaLN) scheme in \cite{adaptive-layernorm} with a Pl\"{u}cker embedding map $\mathbf{p}_{n}$ from its corresponding camera pose parameters \cite{light-field-network}.

This produces a set of pose-aware feature maps $\{\mathbf{f}_n : n=1,\ldots,N\}$.
Then, these modulated feature maps are backprojected into a predefined voxel grid to produce a view-specific feature volume $\mathbf{v}_n \in \mathbb{R}^{H \times W \times D \times C}$, for $n=1,\ldots,N$,
where $H$, $W$, and $D$ are the height, width, and depth of the voxel grid, respectively, and $C$ denotes the number of volumetric feature maps at each view.

\subsubsection{Multi-view feature aggregation \& compression.}
\label{sec:agg}

Next, we propose a lightweight module that aggregates volumetric features from all the input views and transforms it into a compact volumetric representation.
Rather than the computationally expensive transformer architecture \cite{transformer,lara,gs-rgbn}, 
we use a lightweight variant of 3D U-Net \cite{3dunet} after aggregating feature volumes $\{ \mathbf{v}_n : n=1,\ldots,N \}$.

We form an initial unified feature volume $\mathbf{v}^0 \in \mathbb{R}^{H \times W \times D \times 2C}$ as follows:
\begin{equation}
    \mathbf{v}^0 
    = 
    \mathrm{Mean} ( \mathbf{v}_1, \ldots, \mathbf{v}_N )
    \,\mycirc{c}\,
    \mathrm{Var} ( \mathbf{v}_1, \ldots, \mathbf{v}_N ),
\label{eq:v0}
\end{equation}
where $\mycirc{c}$ denotes channel-wise feature concatenation, 
and $\mathrm{Mean}(\cdot)$ and $\mathrm{Var}(\cdot)$ denote the voxel-wise mean and variance operators across views, respectively.
By using the variance operator, we capture voxel-wise deviations in features across views.

To use a lightweight variant of 3D U-Net -- with one encoding block and one decoding block -- for reduced memory and computational cost,
we compress $\mathbf{v}^0$ in (\ref{eq:v0}) into a feature volume with a smaller number of volumetric feature maps $\tilde{C}$ using a 3D convolutional compressor, denoted by $\mathcal{C} : \mathbb{R}^{H \times W \times D \times 2C} \rightarrow \mathbb{R}^{H \times W \times D \times \tilde{C}}$, where $\tilde{C} \ll 2C$. 
The lightweight 3D U-Net variant $\mathcal{U}_\text{light} (\cdot)$ further transforms the compressed feature volume into a unified volumetric representation, denoted $\tilde{\mathbf{v}} \in \mathbb{R}^{H \times W \times D \times \tilde{C}}$.
We formulate this process as follows:
\begin{equation}
\tilde{\mathbf{v}}
=
\mathcal{U}_\text{light} ( \mathcal{C} (\mathbf{v}^0) ).
\label{eq:vol_coarse}
\end{equation}

\subsubsection{Coarse primitive decoder.}
\label{sec:crs-prim}

The final step in the coarse stage is to decode voxel-wise features in $\tilde{\mathbf{v}}$ of (\ref{eq:vol_coarse}) to coarse Gaussian primitives.
We use a fully-connected decoder $\mathcal{D}_\text{crs}(\cdot)$ to decode a feature vector at each voxel $\tilde{\mathbf{v}}_{h,w,d} \in \mathbb{R}^{\tilde{C}}$ into a Gaussian primitive $\mathcal{G}_{h,w,d}^{\text{crs}} = \{ \boldsymbol{\upmu}_{h,w,d}^{\text{crs}}, \mathbf{s}_{h,w,d}^{\text{crs}}, 
\mathbf{r}_{h,w,d}^{\text{crs}}, 
o_{h,w,d}^{\text{crs}}, 
\mathbf{h}_{h,w,d}^{\text{crs}} \}$:
\begin{equation}
\mathcal{G}_{h,w,d}^{\text{crs}} 
= 
\mathcal{D}_\text{crs} ( \tilde{\mathbf{v}}_{h,w,d} ), 
\label{eq:crs-prim}
\end{equation}
for $h = 1,\ldots H, w = 1,\ldots,W$, and $d = 1,\ldots,D$.
As described in the background section, 
we determine $\boldsymbol{\upmu}_{h,w,d}^{\text{crs}}$ by adding a predicted offset $\mathbf{\Delta}_{h,w,d}^{\text{crs}}$ to the coordinate of the $(h,w,d)$th voxel, $\forall h,w,d$.
Note that we predict a single Gaussian primitive per voxel. 

\subsection{The voxel selection stage}
\label{sec:vs}

It is widely known in Gaussian splatting that to achieve high-quality renderings, one needs to position Gaussian primitives near an object \cite{3dgs,2dgs}.
This insight motivates the proposed framework design: 
we first identify voxels that are likely to belong to an object and then predict many Gaussian primitives within those selected voxels.
In this section, we propose a new learnable voxel selection approach that identifies voxels likely to belong to an object, \emph{without} explicit supervised learning using a large-scale 3D geometry dataset \cite{geolrm}.

\subsubsection{Technical motivation of the proposed voxel selection approach.}

A straightforward voxel selection approach is to treat a voxel as belonging to an object to an object if the opacity of its coarse Gaussian, $o_{h,w,d}^{\text{crs}}$ in (\ref{eq:crs-prim}), exceeds a predefined threshold.
However, manually tuning the threshold is challenging. 
More importantly, the opacity alone may \emph{not} be a reliable indicator in determining important Gaussian primitives \cite{lightgaussian}, which, in our framework, corresponds to identifying voxels that are likely to belong to an object.

To circumvent 3D structural supervision in training,
we estimate voxel-wise confidence scores \emph{only} with 2D supervision (i.e., comparing rendered 2D images compared against ground-truth images) using coarse primitives in (\ref{eq:crs-prim}).
In inference, we select voxels with high confidence scores, where one does \emph{not} need to tune a threshold.
We conjecture that selecting important coarse Gaussian primitives -- those associated with high confidence scores -- effectively identifies voxels that are likely to belong to an object.

\subsubsection{Prediction of voxel-wise confidence scores for object membership.}

We propose a learnable module that predicts voxel-wise confidence scores for object membership, $\boldsymbol{\upsigma} \in [0,1]^{H \times W \times D}$.
The score map guides the selection of voxels likely to belong to an object, where fine Gaussian primitives are subsequently placed.

The proposed confidence score predictor consists of a fully-connected predictor $\mathcal{P} : \mathbb{R}^{\tilde{C}} \rightarrow \mathbb{R}$ and a Gumbel-Softmax operator \cite{gumbel-softmax}.
First, for each feature vector $\tilde{\mathbf{v}}_{h,w,d}$ of $\tilde{\mathbf{v}}$ in (\ref{eq:vol_coarse}), 
we predict a confidence logit, i.e., $\mathcal{P}(\tilde{\mathbf{v}}_{h,w,d})$, $\forall h,w,d$.
Next, we model the binary decision for each voxel using a Gumbel-Sigmoid estimator.
To provide a differentiable approximation of discrete binary choices,
we apply a variant of the Gumbel-Softmax distribution, known as the Gumbel-Sigmoid \cite{lp-3dgs}, to confidence logit for each voxel as follows:
\begin{equation}
    \sigma_{h,w,d} = \frac{1}{1+\exp ( -( \mathcal{P}(\tilde{\mathbf{v}}_{h,w,d}) + g_0 - g_1 ) / \tau )},
    \label{eq:conf}
\end{equation}
for $h=1, \ldots H, w=1, \ldots,W$, and $d=1, \ldots D$,
where $g_0$ and $g_1$ denote independent and identically distributed samples drawn from the $\mathrm{Gumbel}(0,1)$ distribution \cite{gumbel}, and $\tau$ denotes the temperature.
Note that Gumbel-Sigmoid-based predictor in (\ref{eq:conf}) stochastically pushes the score toward $0$ or $1$ (especially when $\tau$ is low), so $\upsigma_{h,w,d} \in [0,1]$, $\forall h,w,d$.

To adjust each primitive's contribution in rendered 2D image, 
we propose to refine the opacity of every coarse Gaussian primitive using its confidence score $\sigma_{h,w,d}$ in (\ref{eq:conf}):
\begin{equation}
\label{eq:o}
    \tilde{o}_{h,w,d}^{\text{crs}} 
    = 
    \sigma_{h,w,d} \cdot
    o_{h,w,d}^{\text{crs}},
\end{equation}
for $h=1,\ldots,H, w=1,\ldots W$, and $d=1,\ldots,D$.
In training, 
the proposed scheme (\ref{eq:o}) helps maintain uninterrupted gradient propagation across all voxels.
Using 2D supervision with the proposed refinement (\ref{eq:o}),
we can assign high confidence to coarse primitives that contribute to high-quality rendering.

\subsubsection{Voxel selection.}

\begin{figure}[t!]
    \centering
    \includegraphics[width=1.0\linewidth]{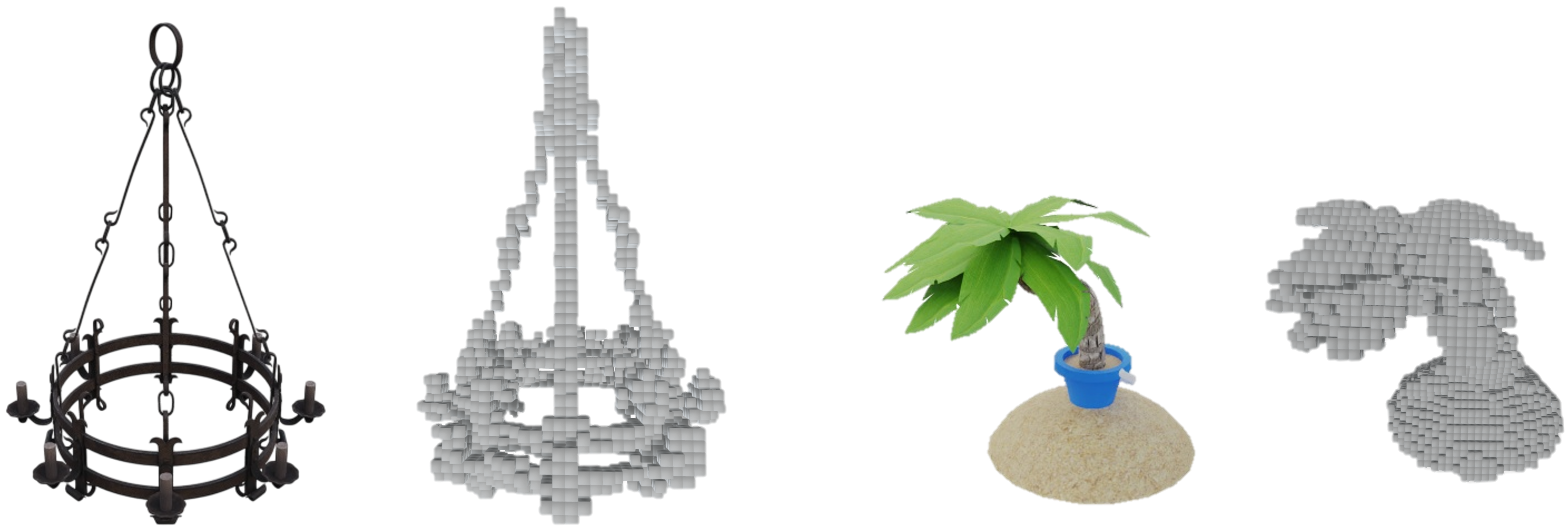}

    \vspace{-0.5pc}
    \caption{ Visualizations of selected voxels with high confidence scores by the proposed learnable voxel selection approach.}
    \vspace{-0.75pc}
    
    \label{fig:vs}
\end{figure}

Now, we select the voxels that are likely to belong to an object if the confidence score in (\ref{eq:o}) exceeds $0.5$,
where we write the skeletal Gaussian primitives by defining the set of the corresponding selected indices, as follows:
\begin{equation}
 \{ \mathcal{G}_{h,w,d}^{\text{crs}} : (h,w,d) \in \mathcal{I} \},
\quad
\mathcal{I} 
= 
\{ (h,w,d) : \sigma_{h,w,d} > 0.5 \},
\label{eq:vs}
\end{equation}
We note that the hard thresholding value of $0.5$ does \emph{not} require tuning, 
as it is consistently used across prior Gumbel-Sigmoid-based works \cite{lp-3dgs,gfs}.
The threshold $0.5$ is a balanced point in the binary decision space.

Ultimately, 
the proposed approach (\ref{eq:conf})--(\ref{eq:vs}) combined with the widely known insight in Gaussian primitive positioning (see the beginning of ``The voxel selection stage'' section)
can promote the selection of voxels that are likely to belong to an object.
Fig.~\ref{fig:vs} supports our conjecture that selected voxels, i.e., coarse Gaussians, with the proposed high confidence scores can capture the 3D structure of an object.

We observed that the proposed voxel selection approach (\ref{eq:conf})--(\ref{eq:vs}) gives higher rendering accuracy, compared to variants of the voxel selection approach motivated by the existing selection methods \cite{GenerativeDensification,lara}. See Appendix~\ref{sec:app:comp} for comparisons with alternative voxel selection approaches.

\subsection{The fine stage}
\label{sec:fs}

The proposed fine stage consists of the following three core modules:
\textit{1)} pose-aware projection at selected voxels,
\textit{2)} selective feature refinement module, and
\textit{3)} fine primitive decoder \& skeletal primitive adjustor.

\subsubsection{Pose-Aware Projection (PAP) at selected voxels.}

The feature volume $\tilde{\mathbf{v}}$ of (\ref{eq:vol_coarse}) in the coarse stage is constructed by backprojecting 2D feature maps from different views to the center of each voxel across the \emph{entire} volumetric space. 
This projection method ignores actual surface geometry and assumes that 2D features can be reliably mapped to the voxel center via camera rays, 
which can lead to ambiguous or inconsistent feature assignment, particularly in occluded or background regions. 
Moreover, associating all features with voxel centers discards sub-voxel geometric precision, potentially limiting the representation of fine-grained details.

To resolve the aforementioned limitations, 
we aim to effectively extract multi-view contextual information \emph{only} for each selected voxel that are likely to belong to an object, where we refer to this process as Pose-Aware Projection (PAP) at selected voxels.

We first project positions of skeletal Gaussians in (\ref{eq:vs}), $\{\boldsymbol{\upmu}^{\text{crs}}_{h,w,d}: (h,w,d) \in \mathcal{I} \}$,
onto each of $N$ pose-aware feature maps $\{ \mathbf{f}_n : n=1,\ldots,N\}$ (obtained in the coarse stage),
and sample the features at the corresponding projected locations.
This results in projected feature volumes tailored to each view, i.e., PAP-derived feature volumes, denoted by $\{ \bar{\mathbf{v}}_{(h,w,d),1}, \ldots, \bar{\mathbf{v}}_{(h,w,d),N}: (h,w,d) \in \mathcal{I} \}$.
By aligning selected 3D skeletal primitives with 2D feature maps from different camera views,
we can allow the network to associate each selected voxel with rich, pose-aligned feature, which would be useful for accurate fine-grained rendering.

We propose to employ inverse distance weighting to aggregate these PAP-derived feature volumes from different views in a voxel-wise manner, 
giving higher importance to those captured from viewpoints closer to the voxel’s primitive location:
\begin{equation}
    \hat{\mathbf{v}}_{h,w,d} 
    = 
    \frac{
    \sum^N_{n=1} z_{(h,w,d),n}^{-1} \cdot \bar{\mathbf{v}}_{(h,w,d),n}
    }
    {
    \sum^N_{n'=1} z_{(h,w,d),n'}^{-1} 
    },
    \quad
    (h,w,d) \in \mathcal{I},
    \label{eq:vol_fine}
\end{equation}
where $z_{(h,w,d),n}$ denotes the distance between $\boldsymbol{\upmu}^{\text{crs}}_{h,w,d}$ and the $n$th view camera position.
{For example, suppose that there are a voxel and three camera views.
The view closest to the voxel will have the greatest influence,
while features from farther views receive lower weights and contribute less.}

\subsubsection{Selective Feature Refinement (SFR) module.}
\label{sec:main:sfr}
While $\tilde{\mathbf{v}}$ in (\ref{eq:vol_coarse}) serves as a globally aggregated coarse representation of the entire 3D object, 
$\{ \hat{\mathbf{v}}_{h,w,d} : (h,w,d) \in \mathcal{I} \}$ in (\ref{eq:vol_fine}) focuses on enriching selected object-centric voxels with fine-grained, pose-aware features derived from 2D views.
With this coarse-to-fine feature hierarchy,
we can capture both global structure and detailed local geometry. 

We propose a new module that refines the coarse, global structure at object-centric voxels, 
$\{ \tilde{\mathbf{v}}_{h,w,d} : (h,w,d) \in \mathcal{I} \}$, 
by integrating fine, view-specific details from the corresponding pose-aware features, 
$\{ \hat{\mathbf{v}}_{h,w,d} : (h,w,d) \in \mathcal{I} \}$.
To provide more accurate spatial location information to each feature $\{ \tilde{\mathbf{v}}_{h,w,d}, \hat{\mathbf{v}}_{h,w,d} \}$, 
we apply positional encoding to the corresponding skeletal primitive position, $\boldsymbol{\upmu}^{\text{crs}}_{h,w,d}$, $\forall (h,w,d) \in \mathcal{I}$.
We refer to this process as Selective Feature Refinement (SFR).

We write its overall process that produces refined object-centric representation $\{ \mathbf{f}^{+}_{h,w,d} : (h,w,d) \in \mathcal{I} \}$ by:
\begin{equation}
\mathbf{f}^{+}_{h,w,d} 
    = 
    \mathcal{R} 
    ( 
    \tilde{\mathbf{v}}_{h,w,d}
    \,\mycirc{c}\,\,
    \hat{\mathbf{v}}_{h,w,d}
    \,\mycirc{c}\,\,
    \gamma(\boldsymbol{\upmu}^{\text{crs}}_{h,w,d})
    ),
    \quad 
    (h,w,d) \in \mathcal{I},
\label{eq:SFR}
\end{equation}
where $\mathcal{R}(\cdot)$ denotes SFR module,
$\tilde{\mathbf{v}}_{h,w,d}$ and $\hat{\mathbf{v}}_{h,w,d}$ are given in (\ref{eq:vol_coarse}) and (\ref{eq:vol_fine}), respectively,
and $\gamma(\cdot)$ is the NeRF positional encoding function \cite{nerf}.
Specifically, we concatenate $\tilde{\mathbf{v}}_{h,w,d}$, $\hat{\mathbf{v}}_{h,w,d}$, and $\gamma(\boldsymbol{\upmu}^{\text{crs}}_{h,w,d})$ in a channel-wise manner at each selected voxel $(h,w,d) \in \mathcal{I}$,
and feed the concatenated results into SFR module.
We then fuse concatenated feature via a fully-connected layer and then process fused feature via $R$ 
residual convolutional blocks \cite{resnet} with two sparse 3D convolution layers.
To further promote intricate details, 
we once more concatenate $\hat{\mathbf{v}}_{h,w,d}$ with proposed fused feature, and 
pass the result through the final sparse 3D convolutional layer.
See the detailed architecture in the Appendix~\ref{sec:app:sfr}.

\subsubsection{Fine primitive decoder \& skeletal primitive adjustor.}

Finally, we generate fine-grained primitives and adjust the skeletal primitives for fine rendering.

We first generate fine-grained primitives ``near'' each skeletal primitive,
where we define a primitive as being near a skeletal primitive if it lies within a $0.5$-voxel radius from the skeletal one.
We use a fully-connected decoder $\mathcal{D}_{\text{fin}} (\cdot)$ to decode refined feature at each selected voxel $\mathbf{f}^{+}_{h,w,d}$ into $M$ fine-grained Gaussian primitives:
\begin{equation}
    \{\mathcal{G}^{\text{fin}}_{(h,w,d),m}: m=1,\ldots,M \} = \mathcal{D}_{\text{fin}}(\mathbf{f}^{+}_{h,w,d}),
    \label{eq:fin-prim}
\end{equation}
where $\mathcal{G}^{\text{fin}}_{(h,w,d),m}$ denotes the $m$th fine-grained Gaussian generated near the $(h,w,d)$th skeletal primitive.
Specifically, we determine the position of each fine-grained primitive, as similarly in (\ref{eq:crs-prim}):
$\boldsymbol{\upmu}^{\text{fin}}_{(h,w,d),m} = \boldsymbol{\upmu}^{\text{crs}}_{h,w,d} + \mathbf{\Delta}^{\text{fin}}_{(h,w,d),m}$, where $\mathbf{\Delta}^{\text{fin}}_{(h,w,d),m}$ denotes the predicted offset, for $(h,w,d) \in \mathcal{I}$ and $m=1,\ldots,M$.
This ensures that fine-grained primitives are near a skeletal primitive at each selected voxel. 

To seamlessly integrate skeletal primitives (\ref{eq:vs}) and fine-grained primitives (\ref{eq:fin-prim}), 
we adjust the appearance parameters -- specifically, opacity and spherical harmonics coefficients -- of skeletal primitives while preserving their structure.
As skeletal primitives represent an object with fewer elements, 
they tend to exhibit higher opacities compared to fine-grained primitives, which may hinder seamlessly integration.
We thus, adjust the appearance parameters of skeletal primitives.

We use a fully-connected adjustor $\mathcal{A}(\cdot)$ to predict the opacity scaling factor $\alpha_{h,w,d}$ and new spherical harmonics coefficients $\mathbf{h}^{\text{new}}_{h,w,d}$ from each refined feature $\mathbf{f}^{+}_{h,w,d}$:
\begin{equation}
    (\alpha_{h,w,d}, {\mathbf{h}^{\text{new}}_{h,w,d}} ) = \mathcal{A}(\mathbf{f}^{+}_{h,w,d}), \quad (h,w,d) \in \mathcal{I}.
\end{equation}
We then update the opacity of skeletal primitives by
\begin{equation}
    o^{\text{upd}}_{h,w,d} = \mathrm{Sigmoid}(\alpha_{h,w,d}) \cdot \tilde{o}^{\text{crs}}_{h,w,d}, 
\quad (h,w,d) \in \mathcal{I},
\end{equation}
where $\tilde{o}^{\text{crs}}_{h,w,d}$ is in (\ref{eq:o}) and $\mathrm{Sigmoid}(\cdot) : \mathbb{R} \rightarrow [0,1]$ denotes the sigmoid function.

Finally, we obtain final rendered images using both find-grained and adjusted skeletal primitives.
During inference, we bypass coarse rendering.

\subsection{Loss function}

We propose to train VS-Splat in an E2E manner using only rendering-based supervision from ground-truth 2D images
$\{\mathbf{i}_n : n = 1,\ldots,N+N' \}$
corresponding to $N$ input viewpoints and $N'$ novel viewpoints, 
\emph{without} explicit 3D structural supervision.
We render the coarse and fine Gaussian representations at the corresponding viewpoints, producing the image sets
$\{\hat{\mathbf{i}}^{\text{crs}}_n: n=1,\ldots,N+N'\}$
and
$\{\hat{\mathbf{i}}^{\text{fin}}_n: n=1,\ldots,N+N'\}$
respectively.
We followed the recent sparse-view setup~\cite{lara,lgm}, $N=N'=4$.
We propose the overall E2E training loss as follows:
\begingroup
\fontsize{9.5pt}{11.4pt}\selectfont
\begin{equation}
    \mathcal{L} 
    := 
    \mathbb{E}_{\mathbf{i}}
    \frac{1}{N+N'}
    \!
    \sum_{n=1}^{N+N'} 
    \!\!
    \sum_{\text{lev} \in \{ \text{crs}, \text{fin} \}} 
    \!
    \mathcal{L}_{\text{MSE}}(\mathbf{i}_n, \hat{\mathbf{i}}^{\text{lev}}_n) +
    \lambda \cdot \mathcal{L}_{\text{SSIM}} (\mathbf{i}_n, \hat{\mathbf{i}}^{\text{lev}}_n),
    \label{eq:loss:base}
\end{equation}
\endgroup
where $\mathbb{E}_{\mathbf{i}}$ denotes the expectation over training object instances,
and $\mathcal{L}_{\text{MSE}}(\cdot,\cdot)$ and $\mathcal{L}_{\text{SSIM}}(\cdot,\cdot)$ denote the mean squared error (MSE) and the structural similarity (SSIM) loss, respectively.
We set the balancing parameter $\lambda$ to $0.5$, following \cite{lara}.

\subsection{Extension for handling noisy camera poses}
\label{sec:ext}

For each spatial feature (e.g., a token embedding) in an image feature map, a Pl\"{u}cker embedding represents the associated camera ray as a six-dimensional vector.
This vector comprises the normalized ray direction $\mathbf{d} \in \mathbb{R}^3$ and the ray moment $\mathbf{m} = \mathbf{o}\times\mathbf{d}$, where $\mathbf{o} \in \mathbb{R}^3$ denotes the camera center and $\times$ denotes the cross product.
For each view $n$, the resulting Pl\"{u}cker embedding map $\mathbf{p}_n$ is transformed into scale and shift parameters that AdaLN~\cite{adaptive-layernorm} uses to modulate the image features and produce a pose-aware feature map (see Section~\ref{sec:fe}).

However, camera poses estimated for in-the-wild datasets such as CO3D~\cite{co3d} may contain noise.
Because each pose determines all camera rays in the corresponding view, these pose errors perturb the Pl\"{u}cker embeddings across the entire feature map.
The base VS-Splat model, trained only with accurate camera poses, is not exposed to such perturbations and thus does not learn to compensate for them.
Consequently, AdaLN modulates the image features using scale and shift parameters derived from the perturbed embeddings, potentially yielding pose-aware feature maps that encode inaccurate camera geometry and are inconsistent across views.

To address this issue, we introduce a SHARE-based extension~\cite{share} that predicts Pl\"{u}cker embedding maps directly from the 2D feature maps, rather than constructing them from potentially noisy camera pose estimates.
Specifically, the extension predicts Pl\"{u}cker embedding maps $\{\hat{\mathbf{p}}_{n} : n=1,\ldots,N\}$ using a two-stage transformer architecture. 
In the first stage, 
a matching transformer applies windowed self-attention within each view and cross-attention across views to establish local feature correspondences between views.
In the second stage, a lightweight transformer jointly processes the correspondence-aware features from all views to capture their global camera configuration.
A regression head then maps each spatial token to a six-dimensional Pl\"{u}cker vector, producing a Pl\"{u}cker embedding map for each view.

We train the Pl\"{u}cker embedding map predictor by supervising its predictions with ground-truth Pl\"{u}cker embedding maps computed from the ground-truth camera poses of input views.
We propose the overall training loss for this extension as follows:
\begin{equation}
    \mathcal{L}_{\text{ext}} 
    := 
    \mathcal{L}
    +
    \mathbb{E}_{\mathbf{i}}\frac{1}{N} \sum_{n=1}^{N}
    \mathcal{L}_{\text{MSE}}([\mathbf{p}(\mathbf{i})]_n, [\hat{\mathbf{p}}(\mathbf{i})]_n),
    \label{eq:loss:ext}
\end{equation}
where $\mathcal{L}$ is defined in (\ref{eq:loss:base}), 
and $[\mathbf{p}(\mathbf{i})]_n$ and
$[\hat{\mathbf{p}}(\mathbf{i})]_n$ denote the ground-truth and predicted Pl\"{u}cker embedding maps, respectively, for the $n$th input view of training object instance $\mathbf{i}$.
Implementation details of the extension are provided in Appendix~\ref{sec:app:exp}.

\begin{table*}[t!]
    \centering
    \caption{Quantitative comparisons of feed-forward novel-view synthesis methods for 3D object reconstruction.\tablefootnote{We obtained the VS-Splat results on GObjaverse and GSO using models trained with the base loss in (\ref{eq:loss:base}), whereas we obtained the CO3D results using the proposed extension trained with the loss in (\ref{eq:loss:ext}). We re-evaluated GeoLRM using its publicly released weights. GeoLRM could not be evaluated on GObjaverse, as its publicly available weights were trained on the full dataset. We followed the experimental setup in \cite{GenerativeDensification}.}}
    \vspace{-0.5pc}
    \setlength{\tabcolsep}{3pt}
    \begin{tabular}{l | ccc | ccc | ccc}
        \toprule
        \multirow{2}{*}{Methods} 
          & \multicolumn{3}{c|}{GObjaverse}
          & \multicolumn{3}{c|}{GSO}
          & \multicolumn{3}{c}{CO3D} \\
        & {PSNR$\uparrow$} & {SSIM$\uparrow$} & {LPIPS$\downarrow$}
        & {PSNR$\uparrow$} & {SSIM$\uparrow$} & {LPIPS$\downarrow$}
        & {PSNR$\uparrow$} & {SSIM$\uparrow$} & {LPIPS$\downarrow$} \\
        \midrule
        MVSNeRF \cite{mvsnerf} & 14.48 & 0.896 & 0.186 & 15.21 & 0.912 & 0.154 & 12.94 & 0.841 & 0.241 \\
        MuRF \cite{murf} & 14.05 & 0.877 & 0.302 & 12.89 & 0.885 & 0.280 & 11.60 & 0.815 & 0.393 \\
        LGM \cite{lgm} &  19.67 & 0.867 & 0.158 & 23.67 & 0.917 & \textbf{0.064} & 13.81 & 0.739 & 0.414 \\ 
        GeoLRM \cite{geolrm} &  - & - & - & 25.06 & 0.947 & 0.068 & 16.49 & 0.829 & 0.284 \\
        LaRa \cite{lara} &  27.51 & 0.938 & 0.093 & 29.81 & 0.959 & 0.071 & 21.18 & 0.862 & 0.216 \\
            \textbf{Ours} &  \textbf{28.77} & \textbf{0.947} & \textbf{0.080} & \textbf{30.84} & \textbf{0.962} & \textbf{0.064} & \textbf{21.62} & \textbf{0.864} & \textbf{0.211} \\
        \midrule
        Lara + GenDen \cite{GenerativeDensification} & 28.58 & 0.945 & 0.080 & 31.06 & 0.966 & 0.058 & 21.72 & 0.865 & 0.209 \\
        \textbf{Ours + GenDen} &  \textbf{29.10} & \textbf{0.949} & \textbf{0.075} & \textbf{31.50} & \textbf{0.968} & \textbf{0.057} & \textbf{22.29} & \textbf{0.867} & \textbf{0.207} \\
        \bottomrule
    \end{tabular}
    \vspace{-0.5pc}
    \label{tab:qr}
\end{table*}
\begin{figure*}[t!]
    \small
    \centering
    \setlength{\tabcolsep}{0.5pt}
    \renewcommand{\arraystretch}{0.0}
    \begin{tabular}{c @{\hspace{3pt}} c c c @{\hspace{4pt}} c@ {\hspace{3pt}} c c}

    \raisebox{1.6\height}{\rotatebox{90}{LGM}} &
    \includegraphics[width=0.17\linewidth]{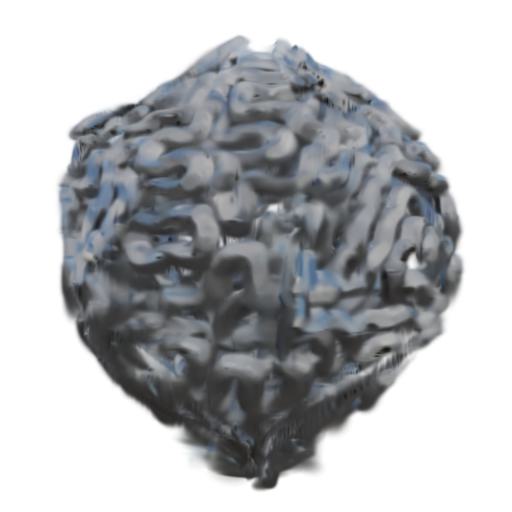} & 
    \includegraphics[width=0.17\linewidth]{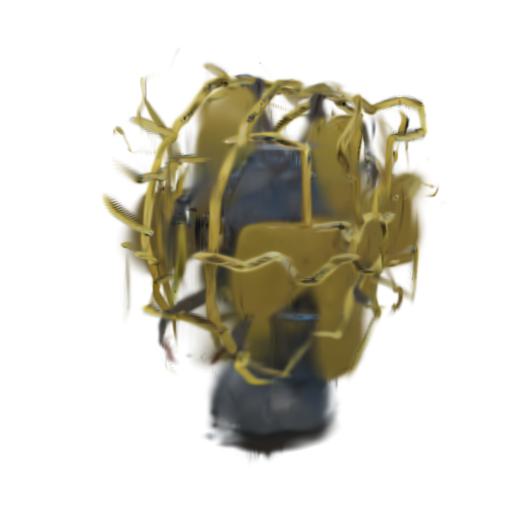} & 
    \includegraphics[width=0.17\linewidth]{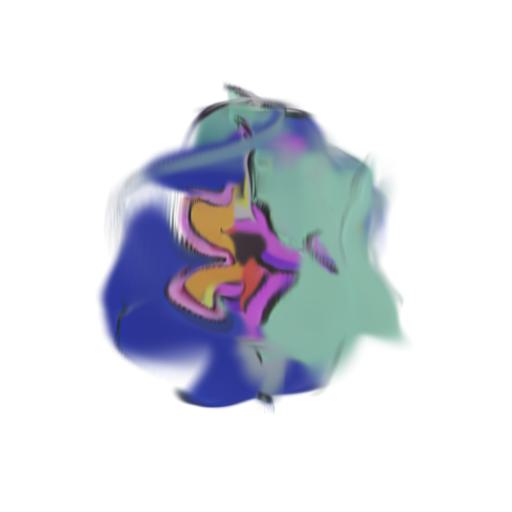} & 
    \raisebox{0.80\height}{\rotatebox{90}{GeoLRM}} &
    \includegraphics[width=0.17\linewidth]{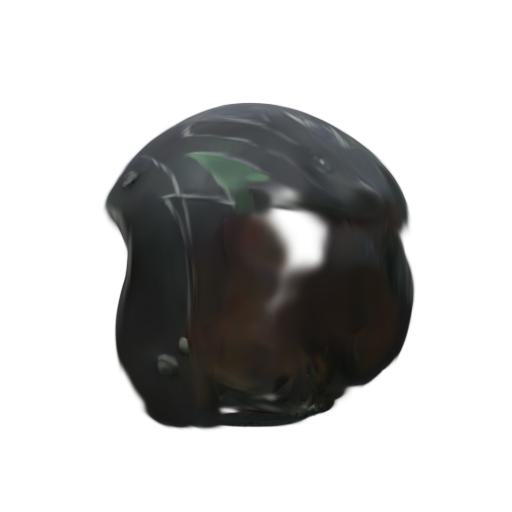} & 
    \includegraphics[width=0.17\linewidth]{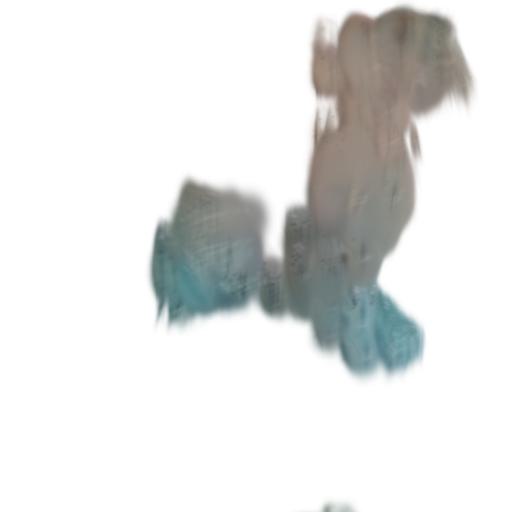} \\[-7pt]

    \raisebox{1.7\height}{\rotatebox{90}{LaRa}} & 
    \includegraphics[width=0.17\linewidth]{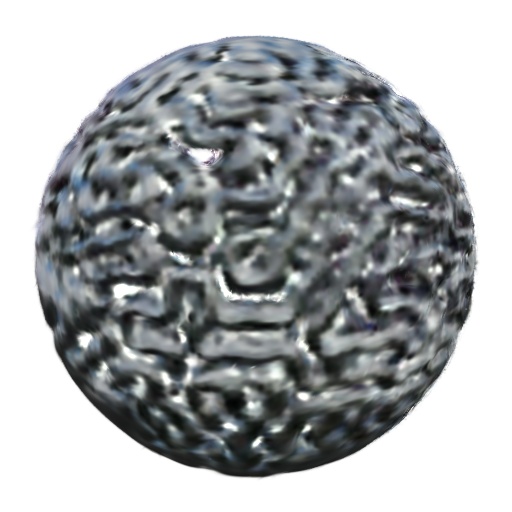} & 
    \includegraphics[width=0.17\linewidth]{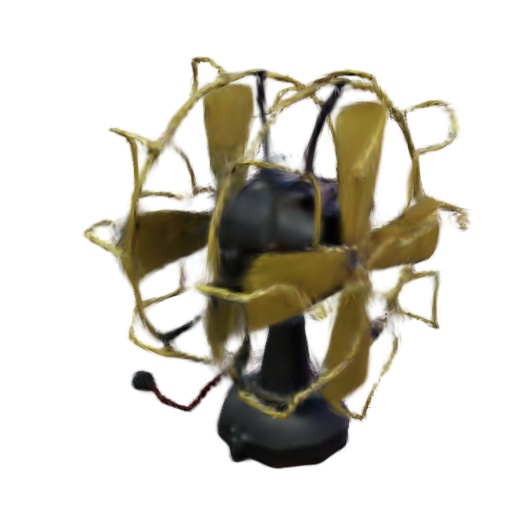} & 
    \includegraphics[width=0.17\linewidth]{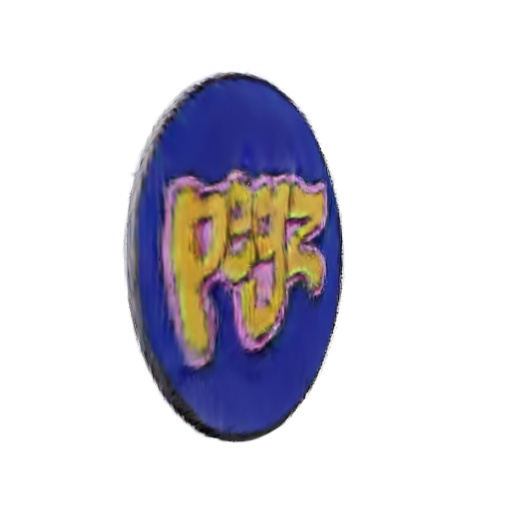} &
    \raisebox{1.7\height}{\rotatebox{90}{LaRa}} & 
    \includegraphics[width=0.17\linewidth]{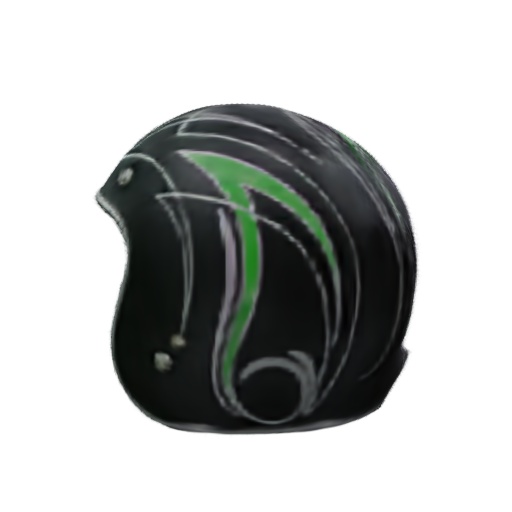} & 
    \includegraphics[width=0.17\linewidth]{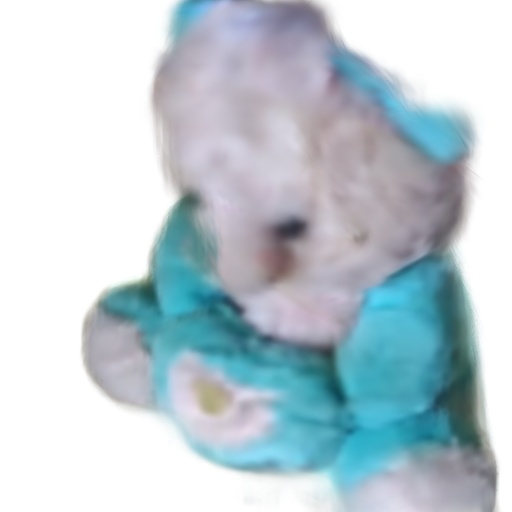} \\[-7pt]

    \raisebox{1.7\height}{\rotatebox{90}{\textbf{Ours}}} & 
    \includegraphics[width=0.17\linewidth]{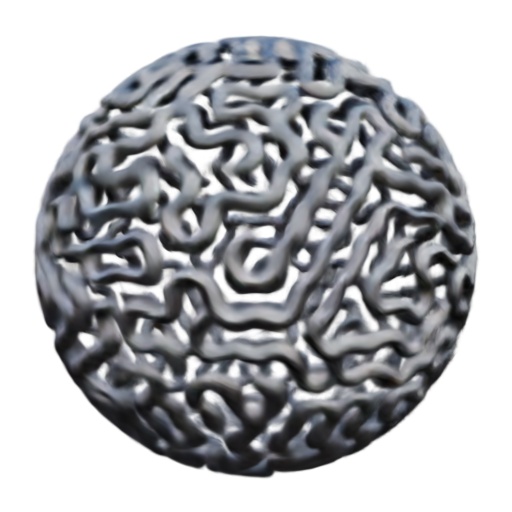} & 
    \includegraphics[width=0.17\linewidth]{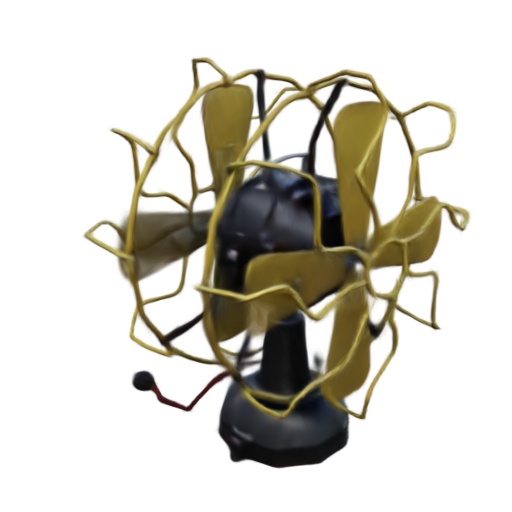} & 
    \includegraphics[width=0.17\linewidth]{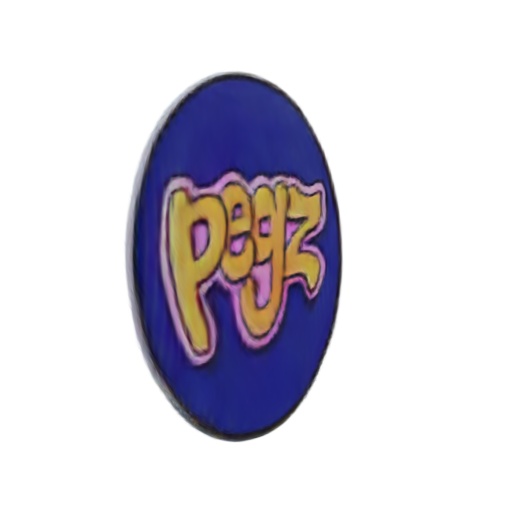} & 
    \raisebox{1.7\height}{\rotatebox{90}{\textbf{Ours}}} & 
    \includegraphics[width=0.17\linewidth]{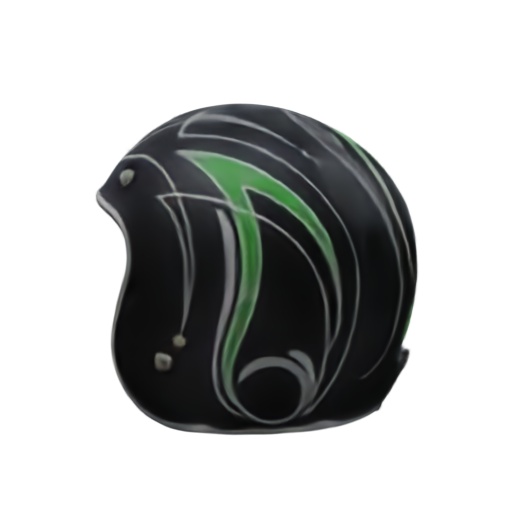} & 
    \includegraphics[width=0.17\linewidth]{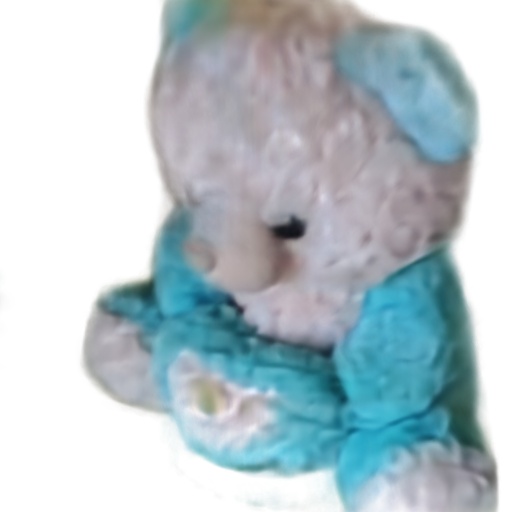} \\[-7pt]

    \raisebox{0.40\height}{\rotatebox{90}{Ground-truth}} & 
    \includegraphics[width=0.17\linewidth]{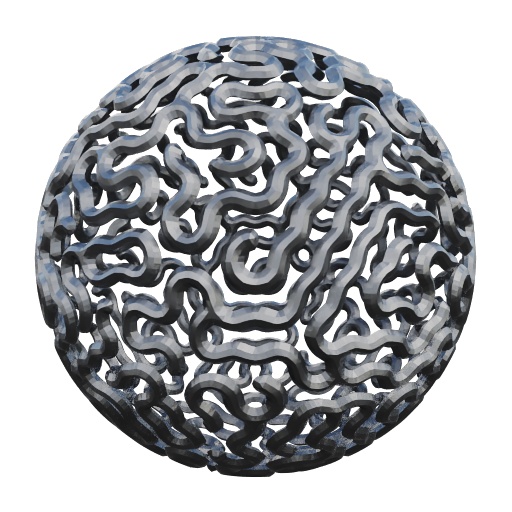} & 
    \includegraphics[width=0.17\linewidth]{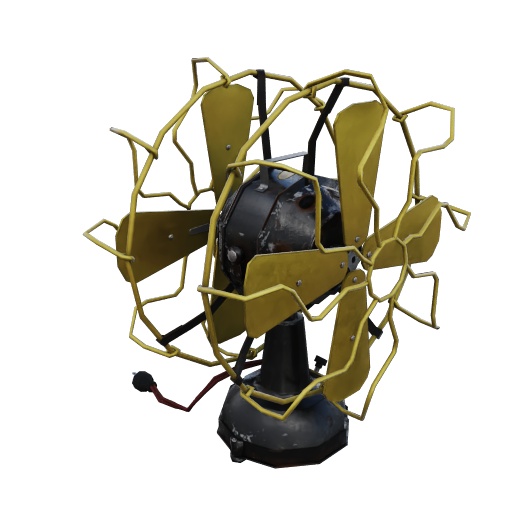} & 
    \includegraphics[width=0.17\linewidth]{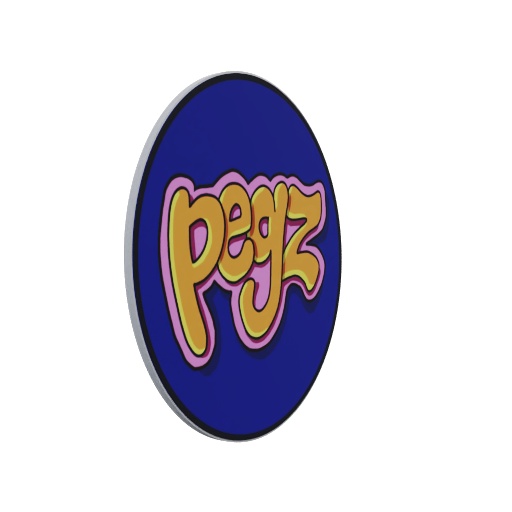} & 
    \raisebox{0.40\height}{\rotatebox{90}{Ground-truth}} & 
    \includegraphics[width=0.17\linewidth]{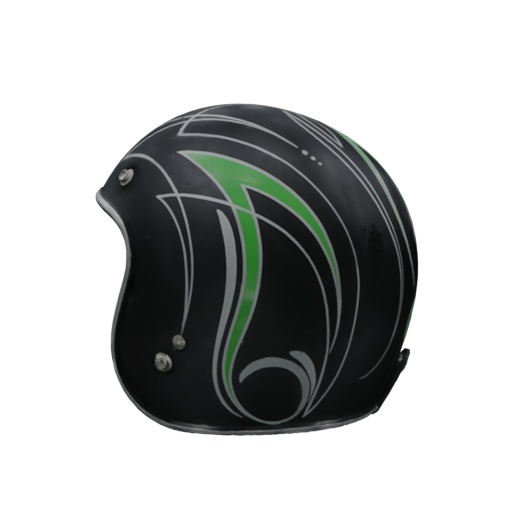} & 
    \includegraphics[width=0.17\linewidth]{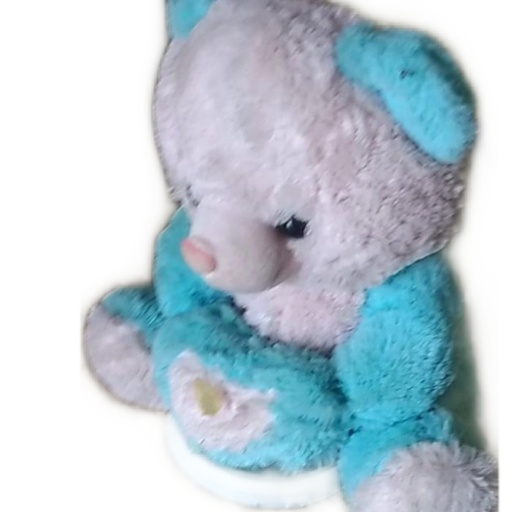} \\ 
    
    & \multicolumn{3}{c}{(a) GObjaverse} & & {(b) GSO} & {(c) CO3D}
    \end{tabular}

    \vspace{-0.25pc}
    \caption{Qualitative comparisons of feed-forward novel-view synthesis methods with different 3D object reconstruction datasets.}
    \vspace{-1pc}
    \label{fig:rg}
\end{figure*}

\begin{table}[t]
    \centering
    \caption{Quantitative comparisons of feed-forward novel-view synthesis methods with GSO (cross-domain setup).\tablefootnote{Since GS-LRM and LVSM do not provide publicly released weights and their papers do not include results on GObjaverse and CO3D, we only report their GSO results. We followed the experimental setup in \cite{gs-lrm}.}}
    \vspace{-0.5pc}
    \setlength{\tabcolsep}{2pt} 
    \begin{tabular}{l | c | cccc}
        \toprule
        Methods
        &  GPUs & {PSNR$\uparrow$} & {SSIM$\uparrow$} & {LPIPS$\downarrow$} & {Time$\downarrow$} \\ 
        \midrule
        \multicolumn{6}{c}{Non Gaussian Splatting} \\
        \midrule
        LVSM (enc--dec) &  8 & 26.48 & 0.901 & 0.065 & 0.78 \\
        LVSM (enc--dec) &  64 & 29.32 & 0.933 & 0.052 & 0.78 \\
        LVSM (dec only) &  8 & 27.04 & 0.910 & 0.055 & 2.56 \\
        LVSM (dec only) &  64 & \textbf{32.36} & 0.962 & \textbf{0.028} & 2.56 \\
        \midrule
        \multicolumn{6}{c}{Gaussian Splatting} \\
        \midrule
        GS-LRM  &  64 & 30.52 & 0.952 & 0.050 & 0.46\\
        \textbf{Ours}  &  4 & 31.14 & 0.965 & 0.061 & \textbf{0.21}\\
        \textbf{Ours+ GenDen} &  4 & 31.91 & \textbf{0.969} & 0.055 & 0.27 \\
        \bottomrule
    \end{tabular}
    \vspace{-1.25pc}
    \label{tab:qr-additional}
\end{table}

\section{Results and discussion}

\subsection{Experimental setups} 

We compared proposed VS-Splat with several SOTA feed-forward novel-view synthesis methods using a few sparse-view images, by following the sparse-view rendering setup in \cite{lara,GenerativeDensification}, $N=4$.
We compared VS-Splat with three categories of SOTA feed-forward methods in 3D object reconstruction:
\textit{1)} GS-based methods, including LaRa \cite{lara}, GS-LRM \cite{gs-lrm}, GeoLRM \cite{geolrm}, and LGM \cite{lgm};
\textit{2)} NeRF-based methods, including MuRF \cite{murf} and MVSNeRF \cite{mvsnerf}; and
\textit{3)} the novel view synthesis method without constructing a renderable 3D representation, LVSM \cite{lvsm}.
The first two categories learn intermediate geometric representations.

We followed the most comprehensive experimental setup \cite{lara,GenerativeDensification} using the three benchmark datasets, large-scale GObjaverse \cite{gobjaverse}, Google Scanned Objects (GSO) \cite{gso}, and  Common Objects in 3D (CO3D) \cite{co3d}. 
As evaluation metrics, we used peak signal-to-noise ratio (PSNR), structural similarity index (SSIM) \cite{ssim}, and learned perceptual image patch similarity (LPIPS) \cite{lpips} as evaluation metrics.
PSNR measures pixel-level rendering fidelity, SSIM evaluates structural similarities including luminance and contrast, and LPIPS assesses perceptual similarity correlated with human visual perception.

\subsection{Comparisons of feed-forward novel-view synthesis methods for 3D object reconstruction}

\begin{table*}[t] 
    \centering

    \caption{Ablation results for VS-Splat (GObjaverse dataset).}
    \vspace{-0.5pc}
    \label{tab:ab}
    \setlength{\tabcolsep}{5pt}
    \begin{tabular}{c|c|c|c|c|c|c|c|c}
        \toprule
        & \multicolumn{5}{c|}{VS-Splat variants} & \multicolumn{3}{c}{Metrics} \\
        \cline{2-9}
        & Coarse stage
        & \specialcell[]{Voxel selection\\ stage} 
        & \specialcell[]{Fine stage: \\ PAP} 
        & \specialcell[]{Fine stage: \\ SFR} 
        & ~~$M$ in (9)~~ 
        & PSNR$\uparrow$ 
        & SSIM$\uparrow$ 
        & LPIPS$\downarrow$ \\
        \midrule
        (a) & $\ocircle$ & $\times$    & $\times$    & $\times$    & N/A & 27.80 & 0.938 & 0.096 \\ 
        (b) & $\ocircle$ & $\ocircle$  & $\times$    & $\times$    & N/A & 27.81 & 0.938 & 0.095 \\
        (c) & $\ocircle$ & $\ocircle$  & $\times$    & $\ocircle$ & 16  & 28.17 & 0.942 & 0.088 \\
        (h) & $\ocircle$ & $\ocircle$  & $\ocircle$ & $\ocircle$ & 16   
            & \textbf{28.40} & \textbf{0.944} & \textbf{0.084} \\
        \bottomrule
    \end{tabular}
    \vspace{-0.25pc}
    \bigskip
    \caption{Quantitative comparison of VS-Splat across different $M$ values (GObjaverse dataset).}
    \vspace{-0.5pc}
    \label{tab:finenum}
    \setlength{\tabcolsep}{5pt}
    \begin{tabular}{c|c|c|c|c|c|c|c|c}
        \toprule
        & \multicolumn{5}{c|}{VS-Splat variants} & \multicolumn{3}{c}{Metrics} \\
        \cline{2-9}
        & Coarse stage 
        & \specialcell[]{Voxel selection \\ stage} 
        & \specialcell[]{Fine stage: \\ PAP} 
        & \specialcell[]{Fine stage: \\ SFR} 
        & ~~$M$ in (9)~~ 
        & PSNR$\uparrow$ 
        & SSIM$\uparrow$ 
        & LPIPS$\downarrow$ \\
        \midrule
        (d) & $\ocircle$ & $\ocircle$ & $\ocircle$ & $\ocircle$ & 1  & 28.28 & 0.942 & 0.087 \\
        (e) & $\ocircle$ & $\ocircle$ & $\ocircle$ & $\ocircle$ & 2  & 28.27 & 0.943 & 0.087 \\
        (f) & $\ocircle$ & $\ocircle$ & $\ocircle$ & $\ocircle$ & 4  & 28.34 & 0.943 & 0.085 \\
        (g) & $\ocircle$ & $\ocircle$ & $\ocircle$ & $\ocircle$ & 8  & 28.34 & \textbf{0.944} & 0.085 \\
        (h) & $\ocircle$ & $\ocircle$ & $\ocircle$ & $\ocircle$ & 16 & \textbf{28.40} & \textbf{0.944} & \textbf{0.084} \\
        \bottomrule
    \end{tabular}
    \vspace{-0.75pc}
\end{table*}

\subsubsection{Comparisons between different GS and NeRF-based feed-forward methods}
\label{sec:main:compff}
Fig.~\ref{fig:rg} and Tables~\ref{tab:qr}--\ref{tab:qr-additional} show that proposed VS-Splat outperforms the existing six SOTA GS- and NeRF-based methods.
Fig.~\ref{fig:rg}(a) shows that VS-Splat produces sharper renderings with finer details compared to SOTA methods.
Fig.~\ref{fig:rg}(b) and \ref{fig:rg}(c) demonstrate superior generalization capability of VS-Splat, consistently achieving significantly better rendering quality compared to SOTA methods across different datasets.
In particular, Fig.~\ref{fig:rg}(c) shows that VS-Splat generalizes well to the real-world CO3D dataset, produces both better overall structure and finer details than other methods.
Tables~\ref{tab:qr}--\ref{tab:qr-additional} show that VS-Splat outperforms the SOTA methods across on all three benchmark datasets.

Now, we compare proposed VS-Splat particularly with the two SOTA GS methods, LaRa and GeoLRM. 
By focusing primitive generation on regions of interest, VS-Splat achieves both accelerated inference and improved results compared to LaRa.
In particular, VS-Splat can achieve up to $1.3\times$ faster inference speed compared to LaRa; see inference speed comparisons with LaRa in Appendix~\ref{sec:app:infspeed}).
While LaRa generates two Gaussian primitives at every voxel, VS-Splat selects on average about $10\%$ of the voxels and generates $17$ primitives (one skeletal and $16$ fine-grained) per selected voxel.
In addition, Fig.~\ref{fig:rg} and Table~\ref{tab:qr} show that VS-Splat outperforms GeoLRM, a multi-stage method that requires 3D geometry supervision and is prone to error propagation.
These results demonstrate that proposed E2E training facilitate high-quality rendering. See Appendix~\ref{sec:app:compext} for comparisons of VS-Splat with and without the extension for handling noisy camera poses.

\subsubsection{Comparisons of augmented feed-forward GS models for 3D object reconstruction}

GenDen is an add-on to LaRa for 3D object reconstruction, further densifying the Gaussian primitives generated by the LaRa backbone~\cite{GenerativeDensification}.
Table~\ref{tab:qr} shows that employing the proposed VS-Splat as the backbone for GenDen (``Ours + GenDen''; see details in Appendix~\ref{sec:app:aug}) consistently outperforms using LaRa as the backbone (``LaRa + GenDen'') \cite{GenerativeDensification} across all datasets.
This result indicates that improvements in a backbone network consistently translate into performance gains when coupled with the add-on model.

\subsubsection{Comparisons with LVSM without constructing a renderable 3D representation}

Table~\ref{tab:qr-additional} shows that, in general, both proposed VS-Splat and its GenDen-augmented variant outperform LVSM, while using substantially fewer GPU resources during training and enabling significantly faster view synthesis.
While slightly underperforming the best-performing LVSM variant, ``LVSM (dec only)'' trained with a large batch size (i.e., using 64 GPUs), the augmented VS-Splat variant achieved approximately $10\times$ faster novel view synthesis and consumes far fewer computational resources during training.
This is because LVSM performs a full network pass for each target view, whereas VS-Splat an explicit 3D Gaussian representation only once per object and renders all target views in real-time.

\subsection{Ablation study}
\label{sec:abl}
To evaluate the contribution of each stage, we conducted an ablation study of VS-Splat with the GObjaverse dataset.
We ran all experiments with learned models using $30$ epochs. We investigated the following VS-Splat variants:
\begin{itemize}
\setlength{\itemsep}{0\baselineskip}
\item (a) only the coarse stage: using coarse primitives in every voxel;
\item (b) the combination of the coarse and voxel selection stages: using skeletal primitives;
\item (c) the combination of the coarse, voxel selection, and fine stages, but excluding the PAP derived features in (\ref{eq:vol_fine});
\item (d)--(g) the full model with $M = 1,2,4,8$ fine-grained primitives per skeletal primitive; and
\item (h) the full model with $M = 16$ fine-grained primitives per skeletal primitive {\bfseries(our default model)}.
\end{itemize}
We did not evaluate the full pipeline with the SFR module removed, 
as this setup is not particularly informative.
Without the SFR module, we simpy use different forms of coarse Gaussian representations, conceptually corresponding to the setup (b).

\subsubsection{Comparisons of rendering perfomance across VS-Splat stages}
This section investigates how different combinations of stages and fine-stage designs affect the performance of VS-Splat.
Comparing (a) and (b) in Table~\ref{tab:ab} shows that using only selected coarse primitives, i.e., skeletal primitives, can achieve rendering quality comparable to using all coarse primitives.
This result implies that the proposed learnable voxel selection approach effectively identifies object-centric voxels that can contribute to high-quality renderings.
Comparing (b) and (h) in Table~\ref{tab:ab} shows that introducing the fine stage with many fine-grained primitives leads to a significant improvement in rendering performance compared to the variant without the fine stage. 
The result suggests that fine-grained primitives serve as an effective representation for modeling object details.

\subsubsection{Comparisons between different design choices in the fine stage}

Next, we study different design choices in the fine stage in terms of the rendering quality. 
Comparing (c) and (h) in Table~\ref{tab:ab} shows that using PAP-derived features in (\ref{eq:vol_coarse}) noticeably improves rendering quality. Comparing the variants from (d) to (h) in Table~\ref{tab:finenum} shows that progressively increasing the number of fine-grained primitives leads to improved rendering fidelity. 
See Appendix~\ref{sec:app:compfine} for qualitative comparisons of these fine-stage design choices.

\section{Conclusion}

In 3D reconstruction and rendering, it is impractical to collect 2D images from many views.
Yet, it is extremely challenging to reconstruct a 3D object from a few sparse-view 2D images.
The proposed VS-Splat, a novel E2E feed-forward GS model, can achieve high-quality renderings from sparse-view images.
Our key idea is to select voxels that are likely to belong to an object and generate many fine-grained Gaussian primitives at those voxels.
To achieve this, we proposed a new \emph{learnable} voxel selection approach that effectively identifies object-centric voxels \emph{without} 3D supervision, and the SFR network that integrates coarse and fine-grained features. 
Beyond its standalone performance, VS-Splat serves as an effective backbone for an existing densification method \cite{GenerativeDensification} and can be extended to improve its robustness to noisy camera pose estimates.

\putbib[main.bib]
\end{bibunit}


\clearpage

\makeatletter
\renewcommand\section{\@startsection{section}{1}{\z@}%
  {3.0ex plus 1.5ex minus 1.5ex}%
  {0.7ex plus 1ex minus 0ex}%
  {\normalfont\normalsize\scshape}}
\makeatother

\renewcommand{\thefigure}{A.\arabic{figure}}
\renewcommand{\thetable}{A.\Roman{table}}
\renewcommand{\thesection}{A.\Roman{section}}
\renewcommand{\theequation}{A.\arabic{equation}}
\renewcommand{\thealgorithm}{A.\arabic{algorithm}} 

\setcounter{section}{0}
\setcounter{equation}{0}
\setcounter{figure}{0}
\setcounter{table}{0}
\setcounter{algorithm}{0}
\setcounter{footnote}{0}

\begin{bibunit}
\twocolumn[
\begin{center}
\fontsize{23.4}{23.4}\selectfont 
VS-Splat: Voxel-Selective feed-forward Gaussian Splatting for end-to-end 3D object reconstruction from sparse-views (Appendix)
\end{center}
]

\maketitle
This appendix consists of seven sections: 
\begin{itemize}
\setlength{\itemsep}{0\baselineskip}
\item Section~\ref{sec:app:comp} compares different voxel selection variants in VS-Splat,
\item Section~\ref{sec:app:sfr} describes the architecture of the Selective Feature Refinement (SFR) module, 
\item Section~\ref{sec:app:exp} describes detailed experimental setup, 
\item Section~\ref{sec:app:infspeed} compares inference speed with LaRa \cite{lara}, 
\item Section~\ref{sec:app:compext} presents comparisons of VS-Splat with and without the extension in Section~\ref{sec:ext} for handling noisy camera poses,
\item Section~\ref{sec:app:aug} presents the implementation details and qualitative results for VS-Splat augmented with GenDen \cite{GenerativeDensification},
\item Section~\ref{sec:app:compfine} presents qualitative comparisons of fine-stage design choices in VS-Splat.
\end{itemize}

\section{Comparisons between different voxel selection variants in proposed VS-Splat}
\label{sec:app:comp}
This section compares different voxel selection variants inspired by existing selection methods, in the proposed VS-Splat framework.
We evaluated them in terms of rendering quality and the average number of selected voxels with the CO3D dataset.
CO3D is a real-world dataset with estimated camera poses, so it is challenging to accurately select object-centric voxels.
We ran all experiments with learned models with $30$ epochs.

We modified the voxel selection variant of VS-Splat with the following selection schemes:
\begin{itemize}
\setlength{\itemsep}{0\baselineskip}
\item (a) opacity-based thresholding (OBT), using the predefined threshold of $0.005$ as in \cite{lara};
\item (b) predicting confidence scores and applying STE to enable gradient flow to the scores, as in \cite{GenerativeDensification};
\item (c) replacing the Gumbel-Sigmoid in (4) with the deterministic Sigmoid function; and
\item (d) our proposed VS-Splat that predicts confidence scores via Gumbel-Sigmoid and uses them to refine opacity values.
\end{itemize}
Note that in (a), we carefully tuned the opacity threshold value (by evaluating inference performances) and observed that $0.005$ yielded the best rendering performance among the candidates $\{ 0, 0.0025, 0.005, 0.0075, 0.01, 0.02, 0.05, 0.1\}$.
Interestingly, excessive generation of fine-grained primitives in non-object (background) regions degraded rendering quality.
We observed this from experiments with the opacity threshold of $0$ that corresponds to the setup of generating many fine-grained primitives across \emph{all} voxels.

\begin{table}[t]
    \caption{Comparisons between different voxel selection variants in the proposed VS-Splat framework (the CO3D dataset).}
    \vspace{-0.5pc}
    \centering
    \setlength{\tabcolsep}{3pt}
    \begin{tabular}{l|cccc}
        \toprule
        Variants & PSNR$\uparrow$ & SSIM$\uparrow$ & LPIPS$\downarrow$ & $\#$ of voxels \\
        \midrule
        (a) OBT & 21.21 & 0.859 & 0.223 & $22,\!867$ \\
        (b) STE & 14.71 & 0.814 & 0.317 & 121 \\
        (c) Sigmoid & 18.01 & 0.835 & 0.288 & $2,\!223$ \\
        (d) $\textbf{Ours}$ & \textbf{21.31} & \textbf{0.863} & \textbf{0.213} & $19,\!587$ \\
        \bottomrule
    \end{tabular}
    \vspace{-0.5pc}
    \label{tab:sel:voxel}
\end{table}

\begin{figure}[t]
    \centering
    \includegraphics[width=1.00\linewidth]{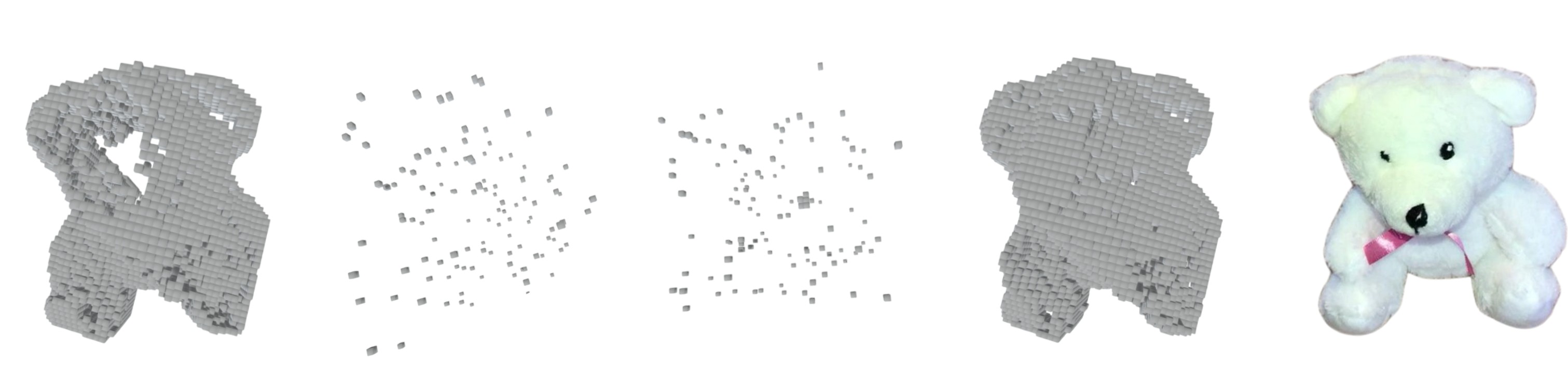}
    \\[-2pt]
    {\footnotesize
    \makebox[0.20\linewidth][c]{(a) OBT}%
    \makebox[0.20\linewidth][c]{(b) STE}%
    \makebox[0.20\linewidth][c]{(c) Sigmoid}%
    \makebox[0.20\linewidth][c]{(d) \textbf{Ours}}%
    \makebox[0.20\linewidth][c]{Ground-truth}%
    }\vspace{-0.5pc}
    \caption{
    Visualizations of selected voxels using different voxel selection variants.
    }
    \vspace{-0.75pc}
    \label{fig:dvs}
\end{figure}

Comparing (a) and (d) in Table~\ref{tab:sel:voxel} demonstrates that the proposed learnable voxel selection approach outperforms the simple OBT scheme across all metrics, while using approximately $15\%$ fewer voxels compared to OBT.
The corresponding visualizations in Fig.~\ref{fig:dvs} show that VS-Splat is significantly better at selecting object-centric voxels compared to OBT.
In particular, Fig.~\ref{fig:dvs}(a) implies that the simple OBT scheme is unreliable in selecting object-centric voxels, corresponding to the argument in \cite{lightgaussian}.
Note that in the OBT scheme, one needs to manually tune a predefined threshold.

Fig.~\ref{fig:dvs}(b) and Table~\ref{tab:sel:voxel}(b) show that simply applying STE to the confidence scores results in selecting an extremely limited number of voxels.
In contrast, our approach enables significantly more appropriate voxel selection and higher-quality renderings over the STE scheme.
This implies that integrating predicted voxel-selection scores into the rendering process (as proposed) is effective in selecting object-centric voxels.
Compare (b) and (d) in Fig.~\ref{fig:dvs} and Table~\ref{tab:sel:voxel}.

Fig.~\ref{fig:dvs}(c) and Table~\ref{tab:sel:voxel} (c) show that using the standard Sigmoid function instead of Gumbel-Sigmoid in (4) fails in properly selecting object-centric voxels, leading to significant rendering quality drops.
Comparing (c) and (d) in Fig.~\ref{fig:dvs} and Table~\ref{tab:sel:voxel} demonstrates that the binary approximation in a stochastic way via Gumbel-Sigmoid encourages the network select object-voxels in an appropriate manner. Fig.~\ref{fig:dvs} and Table~\ref{tab:sel:voxel} show that 
the proposed selection approach better identifies voxels that are likely to belong to an object, compared to the alternative variants derived from existing selection methods.

\section{Architecture of SFR module}
\label{sec:app:sfr}

This section describes the Selective Feature Refinement (SFR) module proposed in (8).
To operate exclusively on selected voxels, 
we build the SFR module with sparse 3D convolutional layers that compute 3D convolution only at non-empty locations \cite{sparsecnn,subm}.
The SFR module consists of fully-connected layer, $R$ residual convolutional blocks with sparse convolution, and a sparse 3D convolutional layer.
Each residual convolutional block consists of two sequential sparse 3D convolutional layers -- each preceded by normalization and Swish activation \cite{swish} -- and has a skip connection.

\begin{figure}[t]
    \centering
    \includegraphics[width=0.60\linewidth]{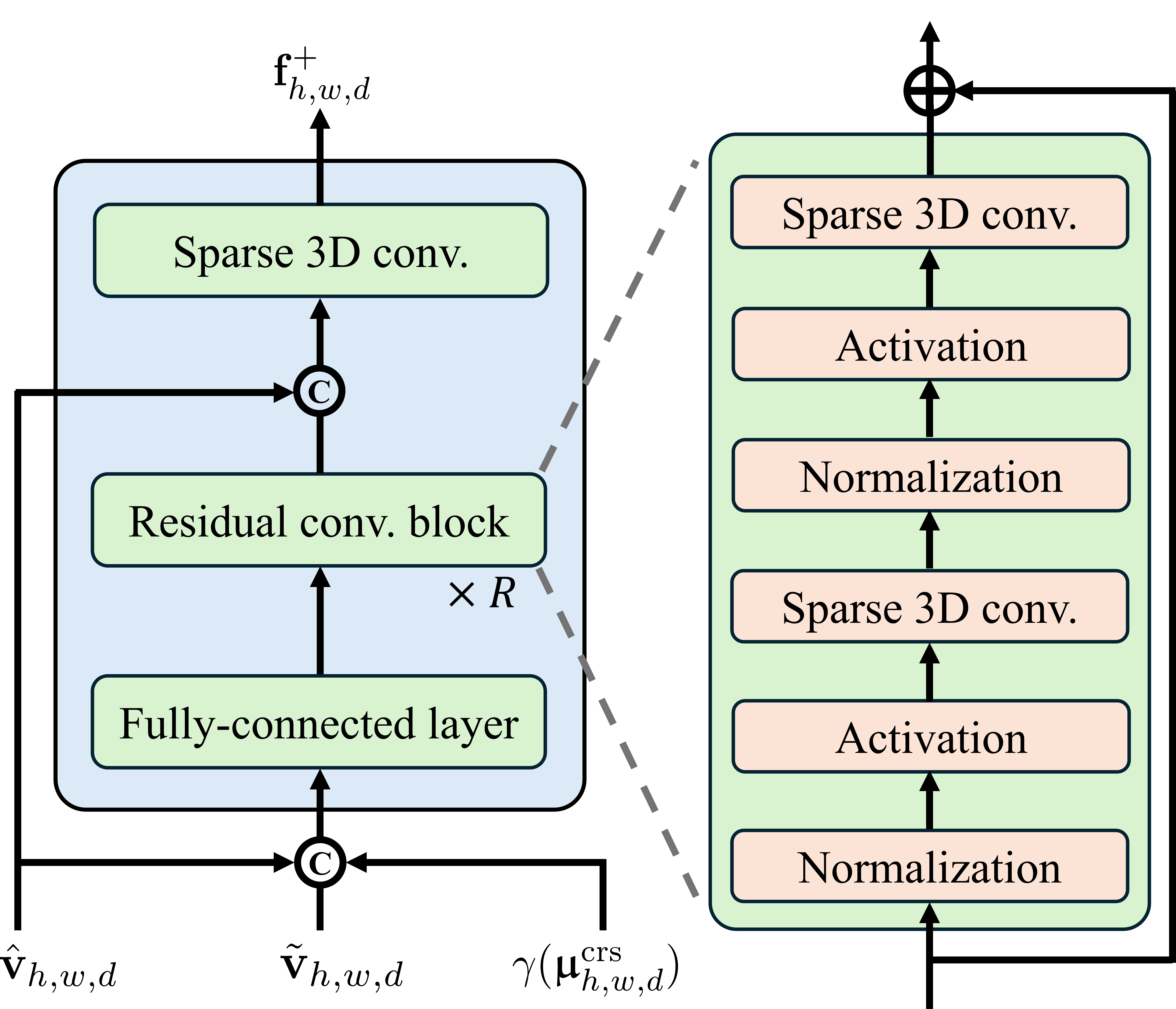}
    \vspace{-0.5pc}
    \caption{
    Architecture of the proposed SFR module.
    ``Sparse 3D conv.'' denotes a sparse 3D convolutional layer, and ``Residual conv.~block'' denotes a residual convolutional block.
    The symbol $\protect\mycirc{c}$ indicates channel-wise concatenation.
    See further details in Section~\ref{sec:app:imp}.}
    \vspace{-0.75pc}
    \label{fig:sfe}
\end{figure}

For each selected voxel $(h,w,d) \in \mathcal{I}$ in (6), 
the SFR takes the unified coarse feature $\tilde{\mathbf{v}}_{h,w,d}$ in (2), 
the fine detail feature $\hat{\mathbf{v}}_{h,w,d}$ in (7), and 
the position-encoded skeletal primitive position $\gamma(\boldsymbol{\upmu}^{\text{crs}}_{h,w,d})$.
The SFR module then concatenates these inputs in a channel-wise manner and ultimately generate the refined object-centric feature $\mathbf{f}^{+}_{h,w,d}$ in (8);
see Section~\ref{sec:main:sfr}.
Figure~\ref{fig:sfe} illustrates the architecture of the proposed SFR module.

\section{Experimental setups}
\label{sec:app:exp}

\subsection{Datasets}

Across all the datasets, 
we used multi-view images of each object with a resolution of $512 \times 512$.
In inference, we input 2D images from four different views, and evaluated each model with rendered 2D images from four novel views.
We trained and evaluated VS-Splat with the large-scale GObjaverse \cite{gobjaverse} dataset, following the preprocessing of \cite{lara}, which comprises $238,\!297$ training and $26,\!478$ evaluation objects.
To assess the cross-domain generalization capability, 
we evaluated VS-Splat with a subset of the Google Scanned Objects (GSO) \cite{gso} and the Common Objects in 3D (CO3D) \cite{co3d}.
GSO comprises $1,\!030$ objects following the preprocessing of \cite{lara}, and CO3D comprises $139$ objects ($47$ hydrants and $92$ teddy bears) following the preprocessing of \cite{GenerativeDensification}.

\subsection{Implementation details}
\label{sec:app:imp}

We adopted the Gaussian representation in 3DGS \cite{3dgs}.
We set the voxel grid resolution to $H=W=D=48$, and generated $M=16$ fine-grained primitives for each selected voxel. 
We linearly annealed the Gumbel-Sigmoid temperature $\tau$ from $1.0$ to $0.1$ over the first $50,\!000$ iterations in training, and fixed it as $0.1$ for the remaining iterations in training, and inference. 

In extracting 2D feature maps from each input 2D images (see Section~\ref{sec:fe}),
we employed DINOv1 \cite{dino}, pre-trained with the ImageNet-1K dataset \cite{imagenet}.
The 3D convolutional compressor $\mathcal{C}$ (see Section~\ref{sec:agg}) reduces the channel dimension from $2C=1,\!536$ to $\tilde{C}=128$.
The SFR module in (8) contains $R=2$ residual convolutional blocks and produces features with a channel dimension of $512$. 
In total, VS-Splat consists of approximately $128$ million trainable parameters.

For the extension in Section~\ref{sec:ext} to handle noisy camera poses, we followed the official implementation of SHARE~\cite{share}, except for a modification to its input module. 
Specifically, we used three $1 \times 1$ 2D convolutional layers to reduce the channel dimension of the feature maps ${\mathbf{i}}_{n=1}^{N}$ extracted in Section~\ref{sec:fe} from $768$ to $128$, matching the input dimension of SHARE. Each convolutional layer is preceded by a normalization layer and a Swish activation~\cite{swish}.
Due to GPU memory constraints, we set $M=8$ for this extension.

We trained VS-Splat and its extension for $50$ epochs with four NVIDIA A100 80GB GPUs, using a batch size of $4$ per GPU.
We used the AdamW optimizer with a learning rate of $4\times10^{-4}$ and applied cosine learning rate scheduling.
We implemented all experiments in Pytorch 2.3.1 \cite{pytorch}, and fixed the random seed to $42$.

\subsection{Details of experimental setups for Table~\ref{tab:qr-additional}}
\label{sec:app:exptab2}
We report only the GSO results of GS-LRM \cite{gs-lrm} and LVSM \cite{lvsm}, since neither method releases public weights nor reports results on GObjaverse and CO3D.
For the proposed VS-Splat, we used DINOv3 pre-trained with LVD-1689M~\cite{dinov3} as the 2D feature extractor.

Following the GS-LRM and LVSM setup, we selected four 2D images from different views, and evaluated our models using rendered 2D images from randomly selected 10 novel views. 
For input view selection, we followed the process in \cite{GenerativeDensification}.
We clustered camera viewpoints into four groups using the classical $K$-means algorithm and selected the cluster centers as input views to ensure an even distribution around the object.

We measured inference speed as the wall-clock time (in seconds) required to process four input images and render 10 novel views.

.

\section{Inference speed comparisons with LaRa}
\label{sec:app:infspeed}

We compared the inference speed of proposed VS-Splat particularly with that of LaRa \cite{lara}, using the GObjaverse, GSO, and CO3D datasets.
For each model, we measured the time taken to process four input images and to generate rendered images for four input viewpoints and four novel viewpoints.
Table~\ref{tab:it} shows that VS-Splat is faster than LaRa across all the benchmark datasets, achieving about a $1.3\times$ speedup with the GObjaverse dataset.
See related discussion of these results in Section~\ref{sec:main:compff}.

\begin{table}[t!]
    \caption{Inference time comparison between LaRa and proposed VS-Splat with three different datasets (in seconds).}
    \vspace{-0.5pc}
    \centering
    \begin{tabular}{l |c c c}
    \toprule
        {Methods} & {GObjaverse} & {GSO} & {CO3D} \\
    \midrule
        {LaRa} & {0.146} & {0.151} & {0.156} \\
        {\textbf{Ours}} & \textbf{0.114} & \textbf{0.139} & \textbf{0.138} \\
    \bottomrule
    \end{tabular}
    \vspace{-0.5pc}
    \label{tab:it}
\end{table}

\begin{table}[t!]
    \caption{Quantitative comparisons of VS-Splat and its GenDen-augmented variant with the extension for handling noisy camera poses (CO3D dataset).\tablefootnote{``Ours" and ``Ours + GenDen" denote VS-Splat and GenDen-augmented VS-Splat, respectively, both trained with the base loss in (\ref{eq:loss:base}). ``Ours$^\text{ext}$" and ``Ours$^\text{ext}$ + GenDen" denote the corresponding models incorporating the extension in Section~\ref{sec:ext}, trained with the extended loss in (\ref{eq:loss:ext}).}}
    \vspace{-0.5pc}
    \centering
    \setlength{\tabcolsep}{3pt}
    \begin{tabular}{l|ccc}
        \toprule
        Variants & PSNR$\uparrow$ & SSIM$\uparrow$ & LPIPS$\downarrow$  \\
        \midrule
        Ours & 21.31 & 0.863 & 0.213 \\
        $\textbf{Ours$^\text{ext}$}$ & \textbf{21.62} & \textbf{0.864} & \textbf{0.211} \\
        \midrule
        Ours + GenDen & 22.09 & 0.865 & 0.208 \\
        $\textbf{Ours$^\text{ext}$ + GenDen}$ & \textbf{22.29} & \textbf{0.867} & \textbf{0.207} \\
        \bottomrule
    \end{tabular}
    \vspace{-0.5pc}
    \label{tab:sel:ext}
\end{table}

\section{Comparisons between with and without the extension for handling noisy camera poses in VS-Splat}
\label{sec:app:compext}

This section compares VS-Splat with and without the extension in Section~\ref{sec:ext} for handling noisy camera poses on the CO3D dataset \cite{co3d}.

A comparison of the results in Tables \ref{tab:qr} and \ref{tab:sel:ext} shows that VS-Splat outperforms the six existing GS- and NeRF-based methods on the CO3D dataset~\cite{co3d}.
This advantage persists when GenDen is applied, as GenDen-augmented VS-Splat (``Ours + GenDen" in Table~\ref{tab:sel:ext}) outperforms ``LaRa + GenDen" in Table~\ref{tab:qr}.
However, the performance gains are smaller on CO3D than on GObjaverse and GSO~\cite{gobjaverse,gso}, likely because CO3D provides less accurate camera pose estimates, whereas GObjaverse and GSO provide accurate camera poses.
Table~\ref{tab:sel:ext} further shows that the extended variants (``Ours$^{\mathrm{ext}}$" and ``Ours$^{\mathrm{ext}}$ + GenDen") outperform their respective counterparts without the extension (``Ours" and ``Ours + GenDen").
These results demonstrate that VS-Splat can be effectively extended to improve its robustness to noisy camera poses.

\begin{figure}[t]
    \centering
    \setlength{\tabcolsep}{6pt}
    \renewcommand{\arraystretch}{1.0}
    {\small
    \begin{tabular}{c c c}
        \includegraphics[
            width=0.22\linewidth
        ]{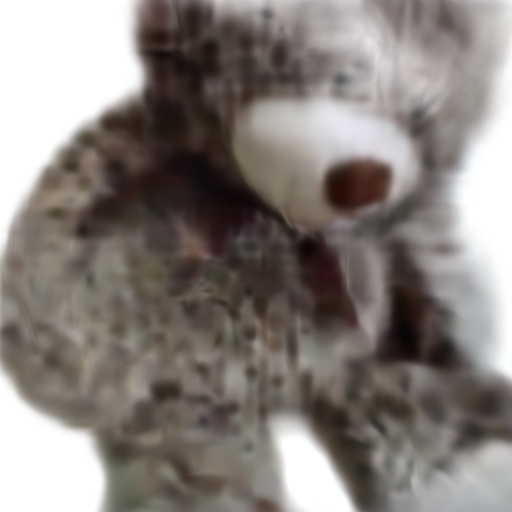} &
        \includegraphics[
            width=0.22\linewidth
        ]{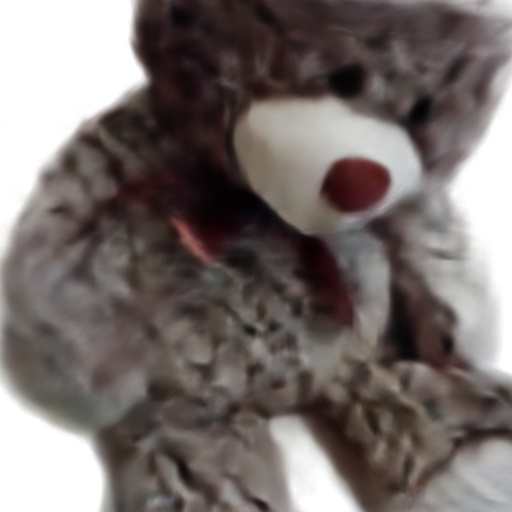} &
        \multirow[t]{4}{*}{
            \hspace{-15pt}\raisebox{-40pt}{
            \begin{minipage}[t]{0.22\linewidth}
                \centering
                \includegraphics[width=\linewidth]{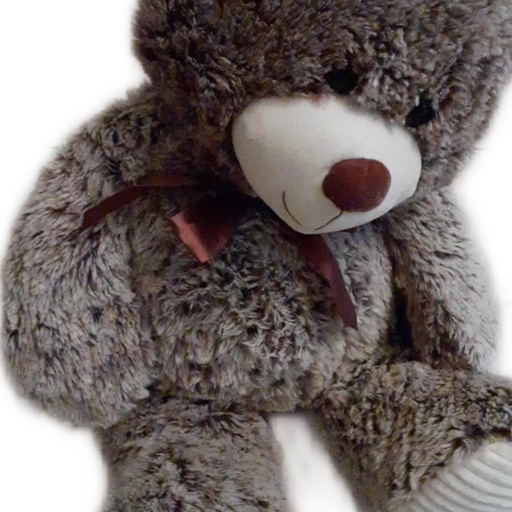}\\[4pt]
                Ground-truth
            \end{minipage}
            }
        } \\[4pt]
        {Ours} & {Ours$^\text{ext}$} & \\[10pt]
        \includegraphics[
            width=0.22\linewidth
        ]{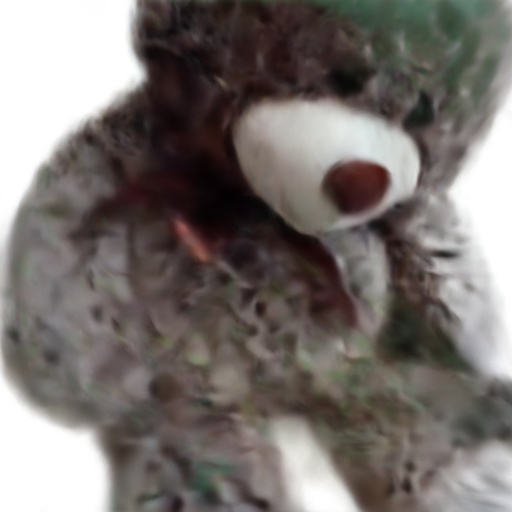} &
        \includegraphics[
            width=0.22\linewidth
        ]{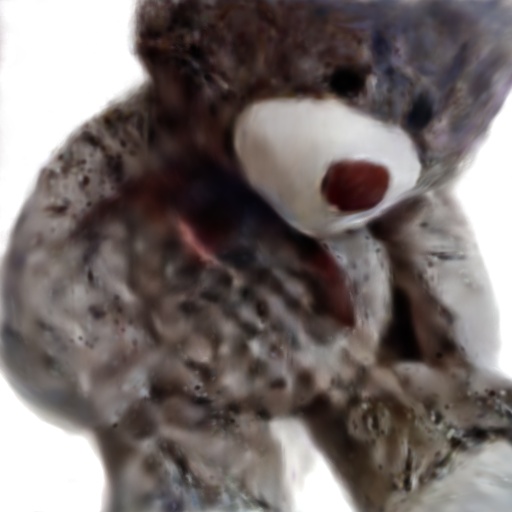} &
        \\[4pt]
        {Ours + GenDen} & {Ours$^\text{ext}$ + GenDen} &
    \end{tabular}
    }
    \caption{Additional qualitative comparisons between VS-Splat and its extension in Section~\ref{sec:ext} for handling noisy camera poses (CO3D dataset).}
    \vspace{-0.75pc}
    \label{fig:posefree}
\end{figure}

Fig.~\ref{fig:posefree} presents qualitative comparisons of renderings produced by VS-Splat and GenDen-augmented VS-Splat, both with and without the extension described in Section~\ref{sec:ext}, on the CO3D dataset~\cite{co3d}.
Consistent with the quantitative results in Table~\ref{tab:sel:ext}, 
the extended models (``Ours$^{\mathrm{ext}}$" and ``Ours$^{\mathrm{ext}}$ + GenDen") produce fewer artifacts and higher-quality renderings than their respective counterparts without the extension (``Ours" and ``Ours + GenDen").

\begin{figure}[t!]
    \centering
    \setlength{\tabcolsep}{2pt}
    \renewcommand{\arraystretch}{0.0}
    \begin{tabular}{c c c}

    \includegraphics[width=0.30\linewidth]{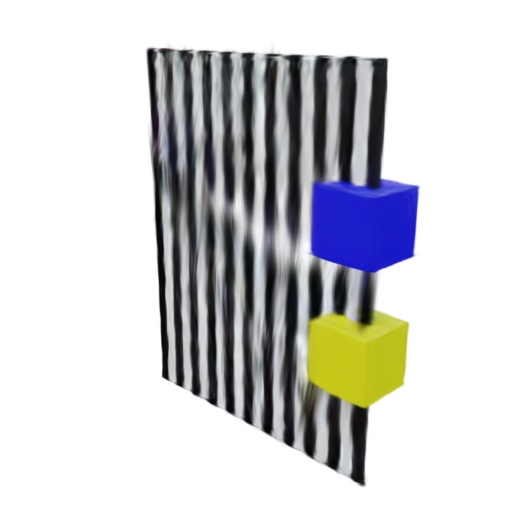} & 
    \includegraphics[width=0.30\linewidth]{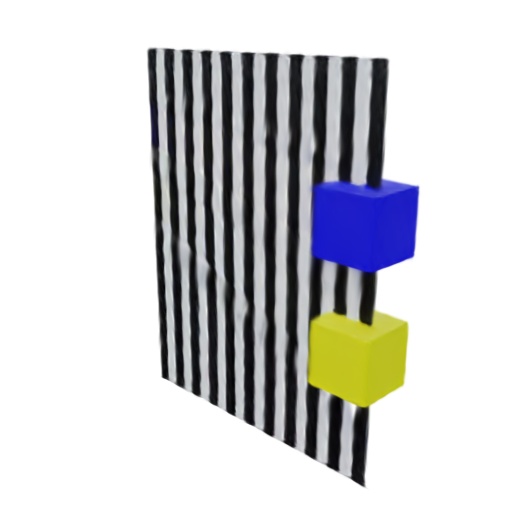} & 
    \includegraphics[width=0.30\linewidth]{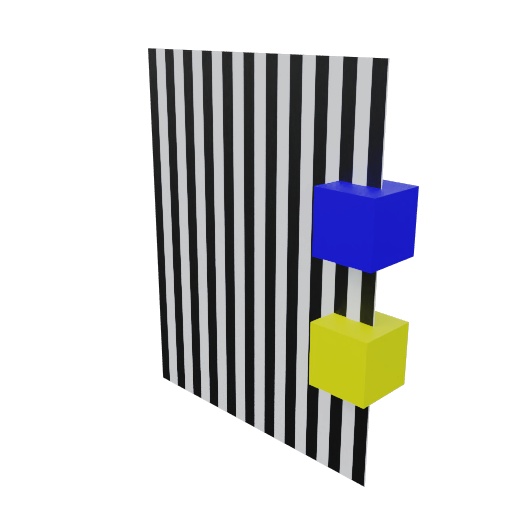} \\[+2pt]
    \includegraphics[width=0.30\linewidth]{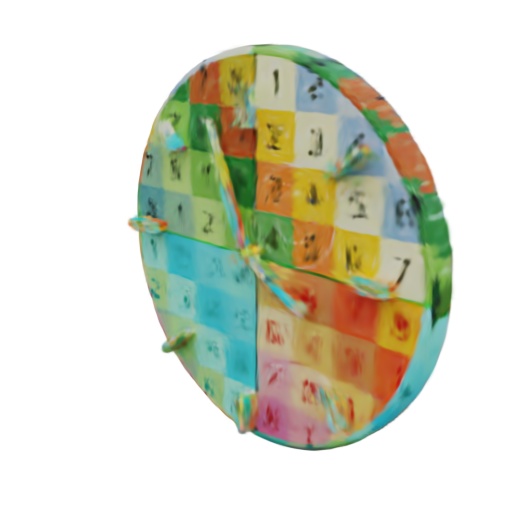} & 
    \includegraphics[width=0.30\linewidth]{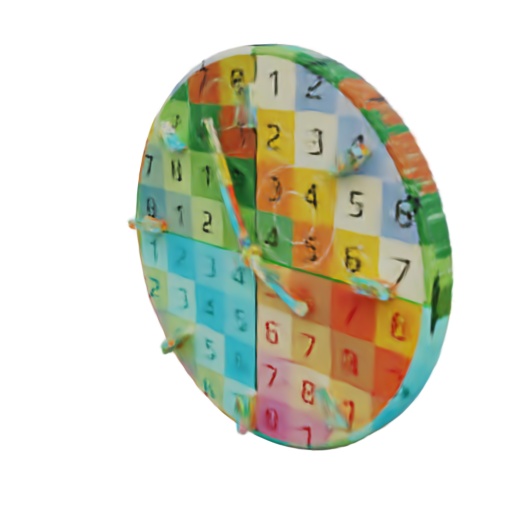} & 
    \includegraphics[width=0.30\linewidth]{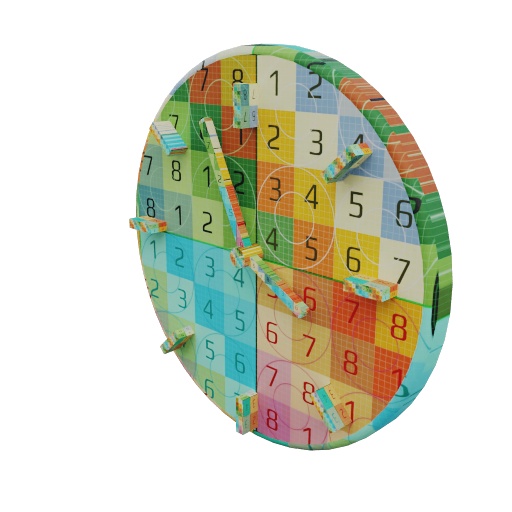} \\[+2pt]
    LaRa + GenDen & \textbf{Ours + GenDen} & Ground-truth 
    \end{tabular}
    \caption{Qualitative comparisons between VS-Splat and LaRa, both augmented with GenDen (GObjaverse dataset).}
    \vspace{-0.5pc}
    \label{fig:genden}
\end{figure}

\begin{figure*}[t!]    
    \centering
    \setlength{\tabcolsep}{6pt}
    \renewcommand{\arraystretch}{1.0}
    {\small
    \begin{tabular}{c c c c}
        \includegraphics[
            width=0.15\linewidth,
            clip,
            trim=0 0 0 4pt
        ]{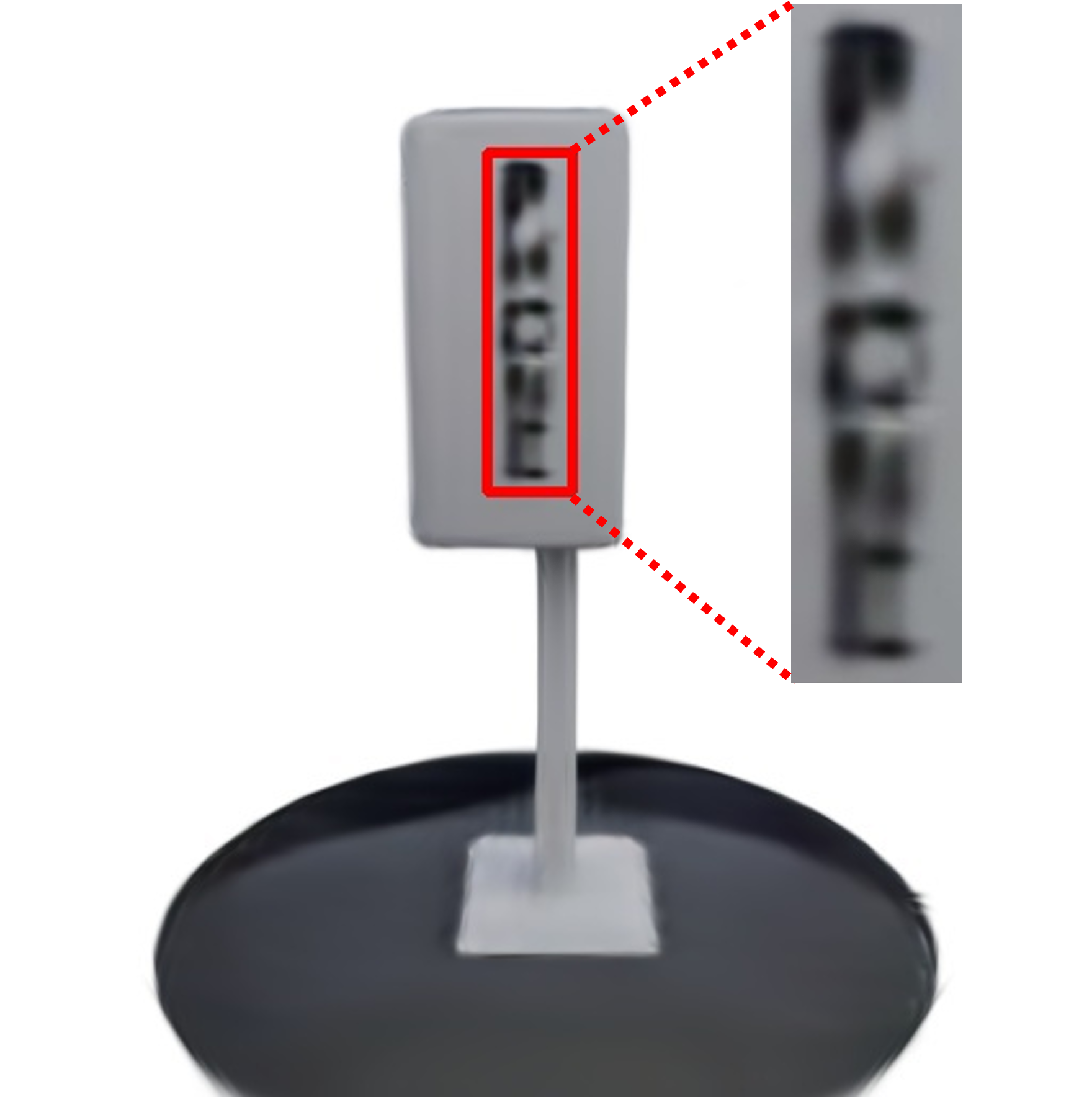} &
        \includegraphics[
            width=0.15\linewidth,
            clip,
            trim=0 0 0 4pt
        ]{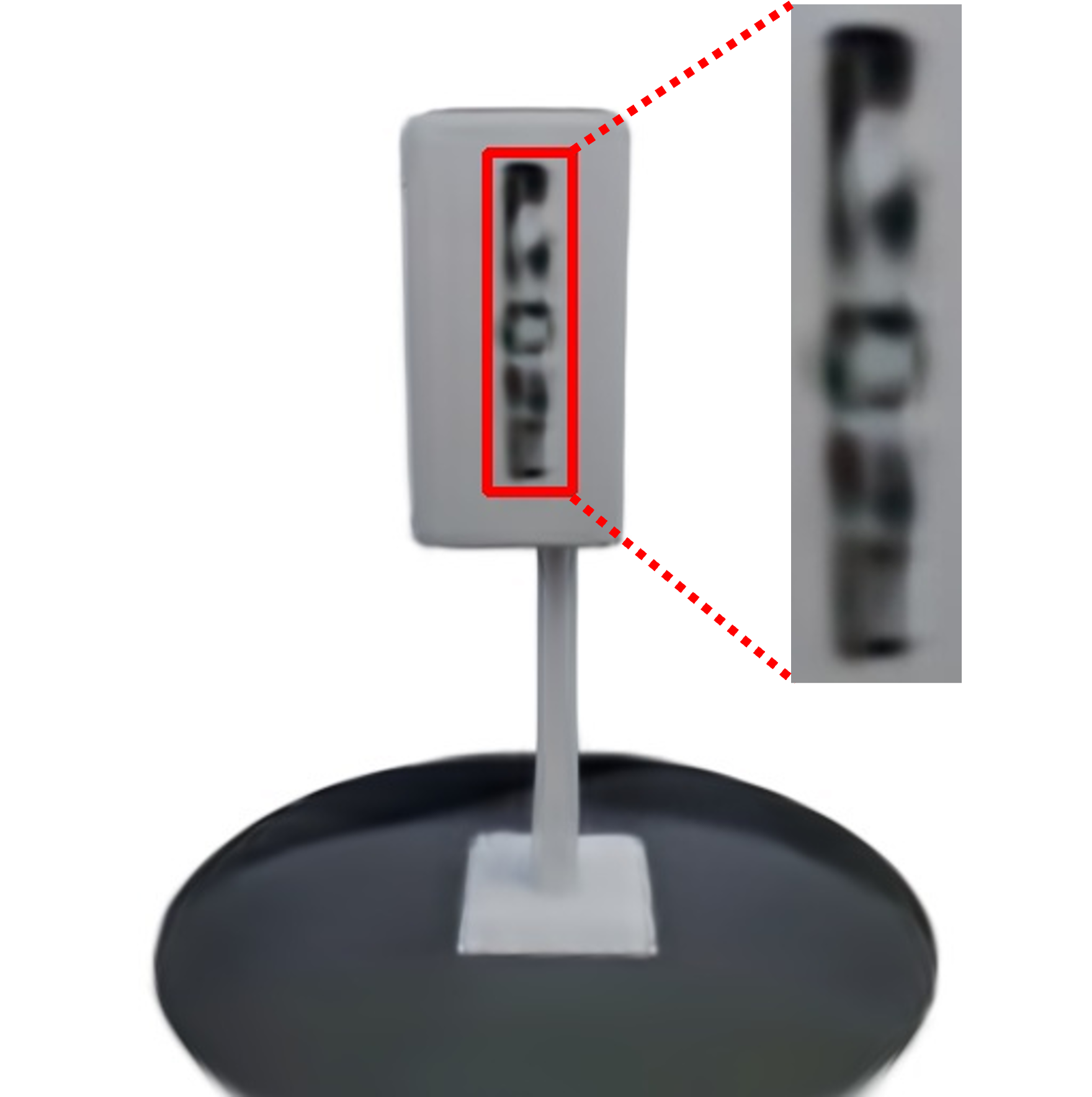} &
        \includegraphics[
            width=0.15\linewidth,
            clip,
            trim=0 0 0 4pt
        ]{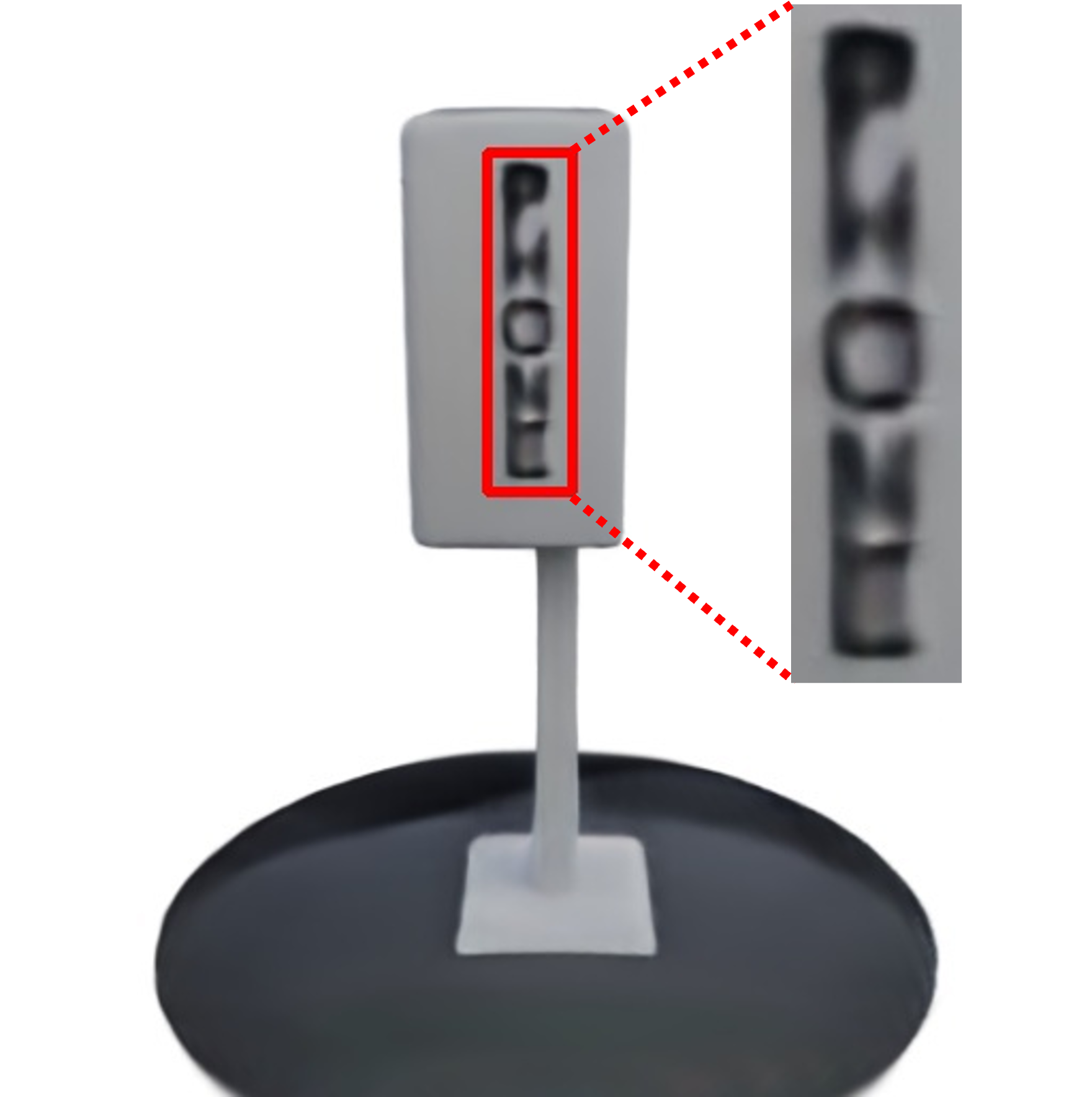} &
        \multirow[t]{4}{*}{ 
            \raisebox{-55pt}{ 
            \begin{minipage}[t]{0.17\linewidth}
                \centering
                \includegraphics[width=\linewidth]{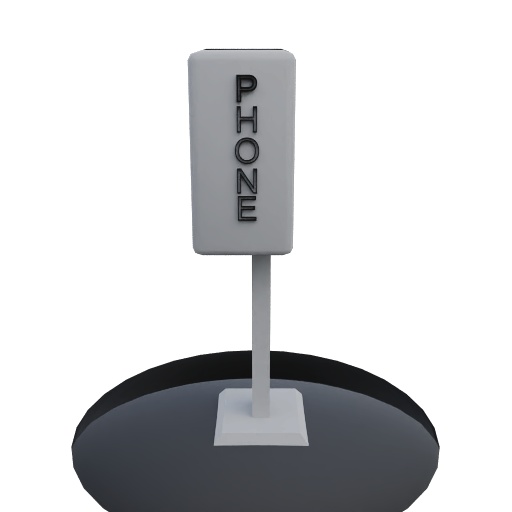}\\ [4pt]
                Ground-truth
            \end{minipage}
            }
        }  \\[4pt] 
        (a) Coarse primitives & (b) Skeletal primitives & \specialcell{(c) W/o PAP driven \\ features ($M=16$)} &  \\[10pt] 

        \includegraphics[
            width=0.15\linewidth,
            clip,
            trim=0 0 0 4pt
        ]{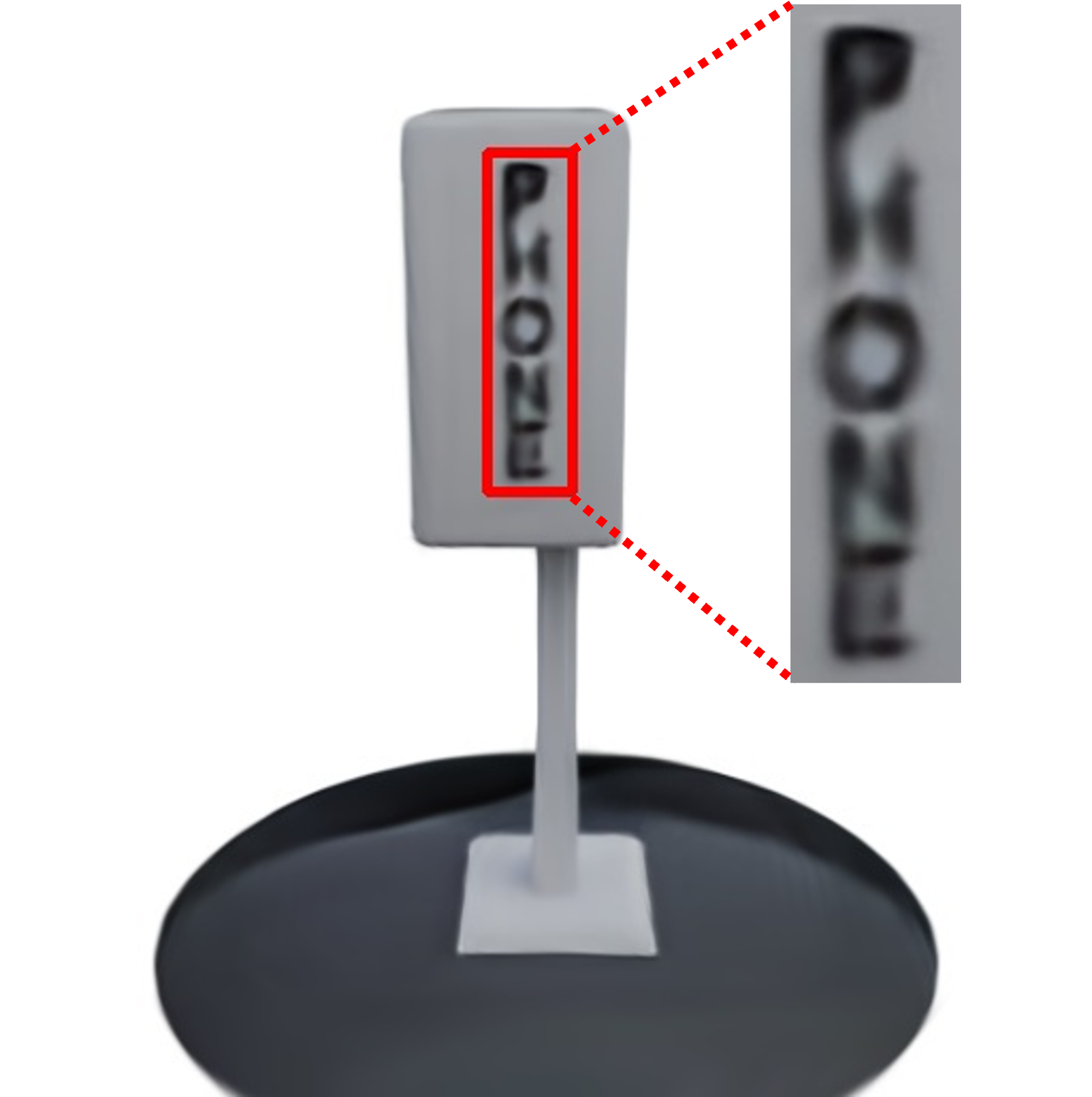} &
        \includegraphics[
            width=0.15\linewidth,
            clip,
            trim=0 0 0 4pt
        ]{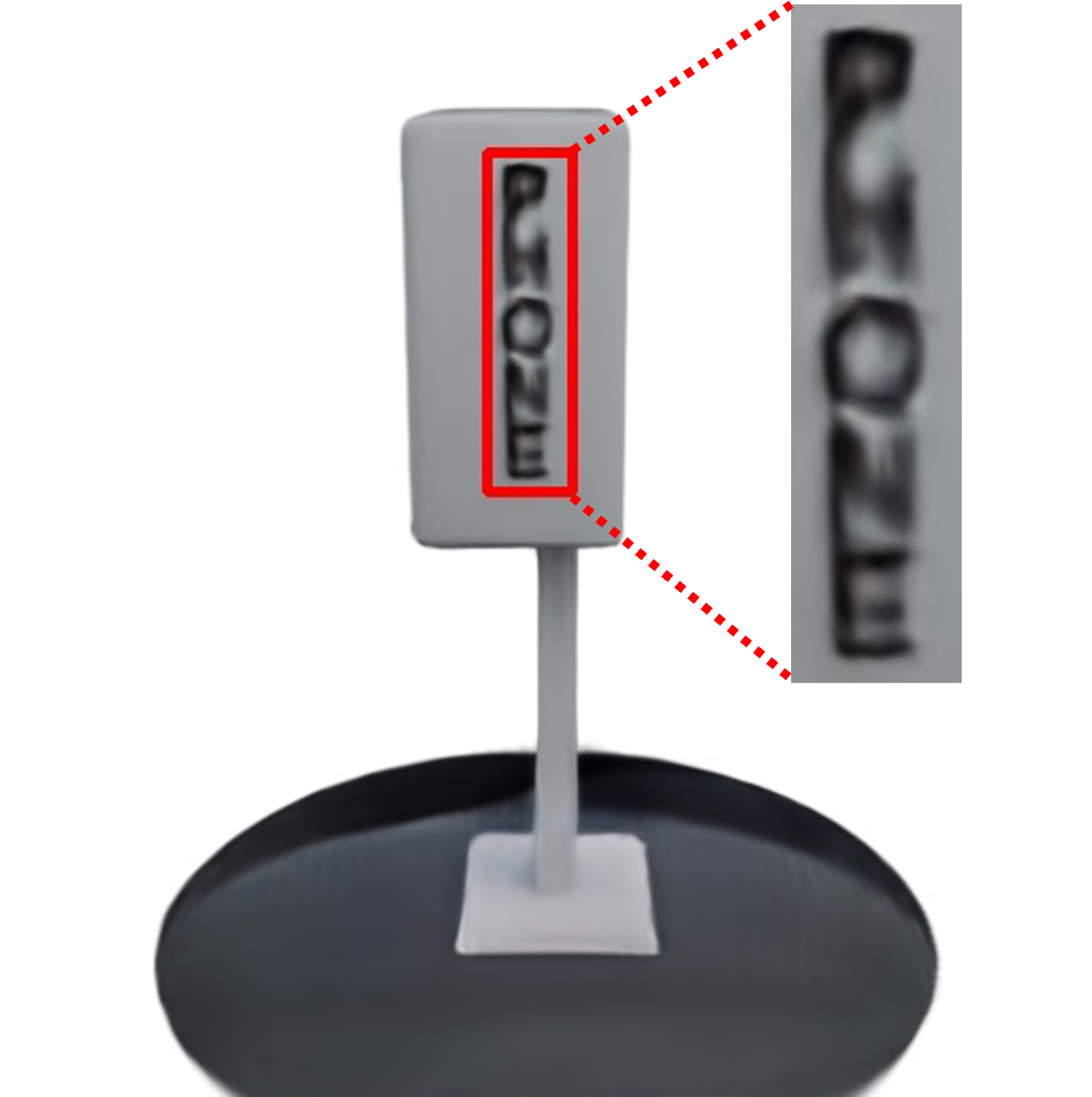} &
        \includegraphics[
            width=0.15\linewidth,
            clip,
            trim=0 0 0 4pt
        ]{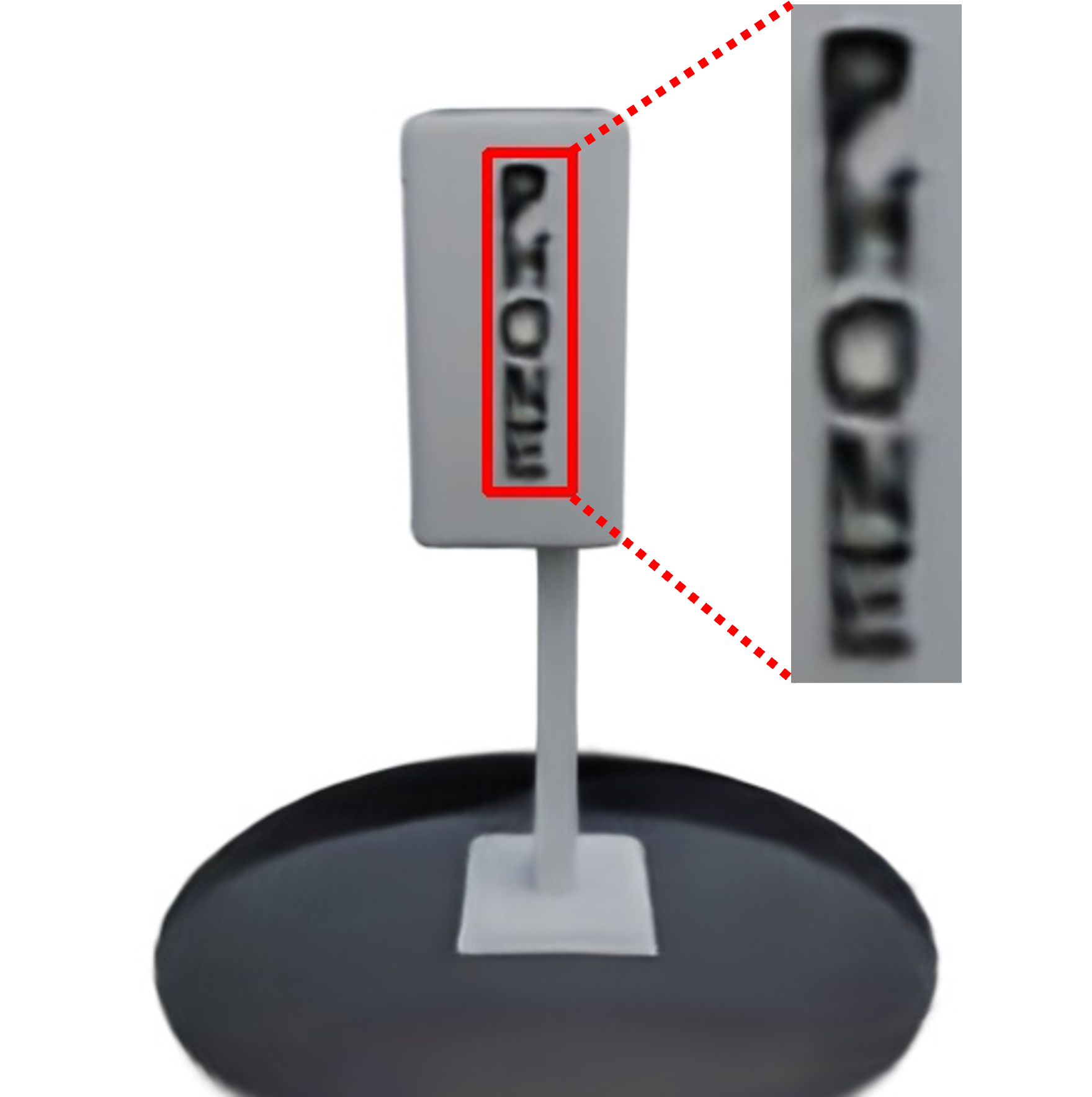} &
        \\[4pt] 
         (d) Full model ($M=1$) & (f) Full model ($M=4$) & \specialcell{(h) Full model \\ ($M=16$; default)} &
         
    \end{tabular}
    }
    \caption{Qualitative comparisons between different VS-Splat variants (GObjaverse dataset). We described the VS-Splat variants (a)-(h) in Section~\ref{sec:abl}.}
    \vspace{-0.75pc}
    \label{fig:ab}
\end{figure*}

\section{Augmenting VS-Splat with an add-on model}
\label{sec:app:aug}

This section describes the implementation details for integrating the GenDen add-on model~\cite{GenerativeDensification} into VS-Splat,
and provides qualitative comparisons of rendered images obtained using VS-Splat and LaRa as backbone models within GenDen.

\subsection{Implementation details}

When using VS-Splat as the backbone, we set $M=8$ rather than $M=16$, due to GPU memory constraints.
All other implementation details are the same as those in Section~\ref{sec:app:imp}.
We followed the official GenDen implementation, replacing the original LaRa backbone with VS-Splat and adjusting input feature dimension of GenDen to match VS-Splat's output feature dimension.
We jointly trained the VS-Splat–GenDen model from scratch for $50$ epochs and evaluated its performance.

\subsection{Qualitative results}
Fig.~\ref{fig:genden} provides qualitative comparisons of rendered images using VS-Splat and LaRa as backbones for GenDen. 
In line with the quantitative results reported in Table~\ref{tab:qr}, Fig.~\ref{fig:genden} demonstrates that GenDen achieves better rendering quality when built on VS-Splat rather than LaRa.
This result demonstrates that VS-Splat can serve as an effective backbone for the GenDen densification method.

\section{Qualitative comparisons of fine-stage design choices}
\label{sec:app:compfine}

Fig.~\ref{fig:ab} presents qualitative comparisons that complement the quantitative results in Tables~\ref{tab:ab} and~\ref{tab:finenum}. 
A comparison of panels Fig.~\ref{fig:ab}(c) and Fig.~\ref{fig:ab}(h) shows that incorporating the PAP-derived features $\{ \hat{\mathbf{v}}_{h,w,d}: (h,w,d) \in \mathcal{I}  \}$ in (\ref{eq:vol_fine}) improves rendering quality, 
with Fig.~\ref{fig:ab}(h) reconstructing fine local details more accurately than Fig.~\ref{fig:ab}(c). 
This result indicates that the PAP-derived features provide useful fine-grained information for high-fidelity rendering.

Fig.~\ref{fig:ab}(d)--(h) further shows that increasing the number of fine-grained Gaussian primitives ($M$) progressively improves the clarity of fine details, consistent with the quantitative results in Table~\ref{tab:finenum}.

\putbib[appendix.bib]
\end{bibunit}

\end{document}